%% file: main_arxiv.tex
\documentclass[10pt,twocolumn,letterpaper]{article}

\usepackage[pagenumbers]{cvpr} 

\input{preamble}
\usepackage{float}
\usepackage{marvosym}

\definecolor{cvprblue}{rgb}{0.21,0.49,0.74}
\usepackage[pagebackref,breaklinks,colorlinks,allcolors=cvprblue]{hyperref}

\def\paperID{*****} 
\def\confName{arXiv}
\def\confYear{2026}

\title{Copper-Policy: Focus on the Representation for Robust Robot Manipulation}

\author{
    Zexin Feng\textsuperscript{1} \quad Yixu Feng\textsuperscript{2} \quad Lingyu Xiao\textsuperscript{1} \quad Shang Su\textsuperscript{3,4} \quad Kexin Zheng\textsuperscript{1} \\
    Chang Xu\textsuperscript{2} \quad Mengkai Shi\textsuperscript{4} \quad Shuo Feng\textsuperscript{3} \quad Xintao Yan\textsuperscript{1,\Letter} \\[5pt]
    \textsuperscript{1}The University of Hong Kong \quad
    \textsuperscript{2}The University of Sydney \\ 
    \textsuperscript{3}Tsinghua University \quad
    \textsuperscript{4}DenseAI Inc.
}

\begin{document}
\twocolumn[{%
    \renewcommand\twocolumn[1][]{#1}%
    \maketitle
    \begin{center}
        \centering
        \includegraphics[width=0.99\textwidth]{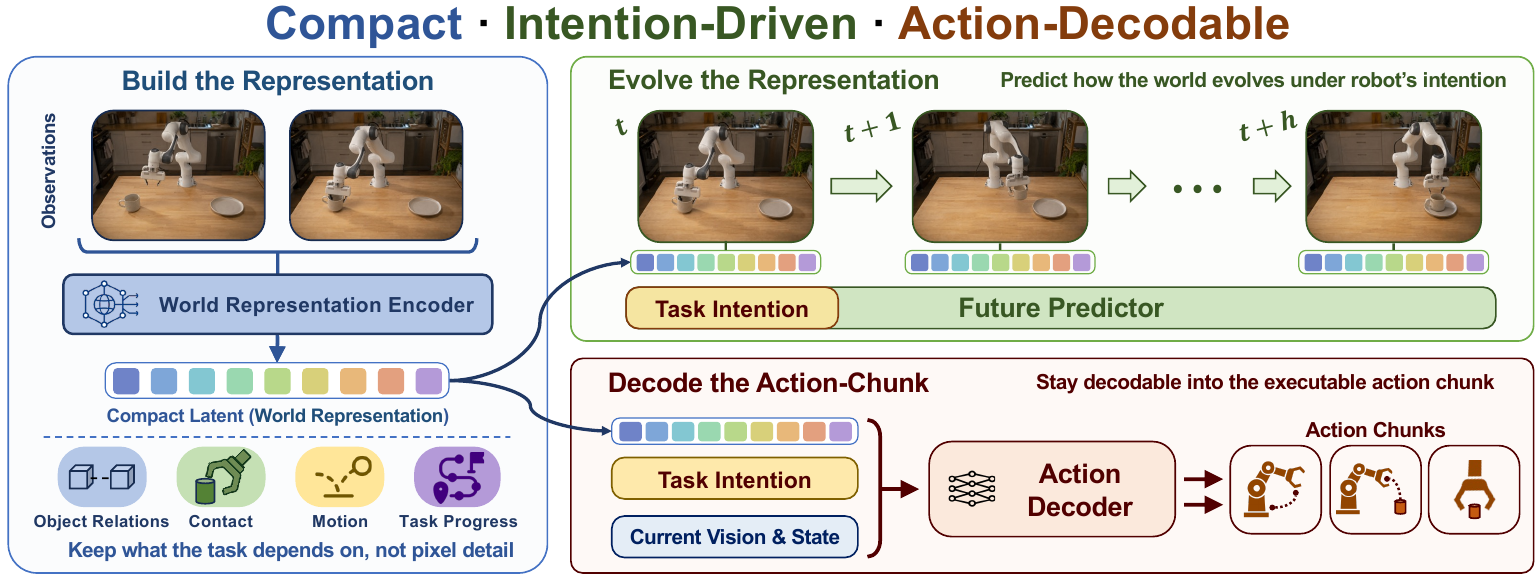}
        \captionof{figure}{\textbf{Overview of our formulation.} A World representation for control should be compact, evolve under task intention, and remain decodable into actions.}
        \label{fig:teaser}
    \end{center}
    \vspace{1em}
}]
\begingroup
\renewcommand{\thefootnote}{\Letter}
\footnotetext{Corresponding author.}
\endgroup

\input{sec_ARXIV/0_abstract}
\input{sec_ARXIV/1_intro}
\input{sec_ARXIV/2_related_works}
\input{sec_ARXIV/4_method}
\input{sec_ARXIV/5_experiments}

\input{sec_ARXIV/6_conclusion}
\clearpage
{
    \small
    \bibliographystyle{ieeenat_fullname}
    \bibliography{main}
}

\onecolumn
\appendix
\setcounter{table}{0}
\setcounter{figure}{0}
\renewcommand{\thetable}{A\arabic{table}}
\renewcommand{\thefigure}{A\arabic{figure}}
\input{sec_ARXIV/X_suppl}

\end{document}

%% file: preamble.tex
\usepackage{multirow}

\newcommand{\supmark}[1]{\raisebox{0pt}[0pt][0pt]{\ensuremath{^{#1}}}}

%% file: sec_ARXIV/0_abstract.tex
\begin{abstract}
World Action Models (WAMs) acquire behavioral priors by modeling future scene evolution, but predicting detailed futures in pixel or latent space incurs substantial cost. Recent evidence that co-training gains persist without test-time generation raises a question: what must a WAM learn to improve control? We introduce \textbf{\emph{Copper-Policy}}, which learns a compact World representation with the policy rather than relying on a predefined target space. Through temporal joint-embedding prediction, it predicts future observation embeddings conditioned on task intention without reconstructing pixels. This prediction and action decoding shape the representation jointly, while the policy retains access to current-frame spatial detail for execution. Representation analyses show that the learned features better separate task-driven change from perturbations and provide complementary information for control. Compact prediction targets reduce training tokens per sample, enabling a 2B-parameter model trained in 9.67 hours on 8$\times$ RTX 5090 GPUs, about 6$\times$ faster than Fast-WAM. \emph{Copper-Policy} leads all compared methods without embodied pretraining on RoboTwin and outperforms several embodied-pretrained VLAs on LIBERO-Plus (80.85\%). On three real-robot tasks, its 96.3\% average is comparable to $\pi_{0.5}$. Together, these results support strong control performance and efficient training.
\end{abstract}

%% file: sec_ARXIV/1_intro.tex
\section{Introduction}
\label{sec:intro}

Robotic research has long pursued autonomous systems equipped with human-like cognition to enable daily physical interaction. This pursuit has been accelerated by recent advances in the concept of \textbf{\emph{world modeling}}, which offers a shared substrate for building decision-oriented cognition that can both \emph{comprehend} and \emph{predict what will happen}, specialized for downstream embodiment control. One prominent backbone is the Vision-Language-Action (VLA) model, which capitalizes on the semantic capabilities of pretrained Vision-Language Models (VLMs) to establish a direct mapping from visual observations and language instructions to executable robot actions~\citep{pmlr-v270-kim25c,intelligence2025pi_,zheng2026x,yang2026abot}. In contrast, the emerging line, World Action Models (WAMs), go further by modeling future scene evolution through visual dynamics~\citep{yuan2026fastwam,li2026causal}, explicitly linking \textit{what the world should become} to \textit{how to achieve it}. Yet despite their effectiveness, generative WAMs couple action with explicit video (or latent video) generation, leading to substantial training burden and inference overhead.

This limitation necessitates a reassessment: \textbf{\emph{what aspect of dynamics learning in WAM actually drives better control?}} Fast-WAM~\citep{yuan2026fastwam} demonstrates that video co-training remains beneficial even without test-time future generation, while more recent studies indicate that the learned embeddings that encode informative future representations could actively shape policy decisions~\citep{pmlr-v305-zheng25a,su2026world,beingbeyond2026beingh07}. The value of WAMs thus appears to lie less in the futures they generate than in the representations their training shapes, motivating a latent-prediction paradigm that predicts future states directly in representation space. Existing designs largely fall into two categories. One line uses a pretrained Joint-Embedding Predictive Architecture (JEPA) space either as additional policy input or as fixed prediction targets~\citep{sun2026vla,lin2026jepawamlearningvisionlanguageactionpolicies}. The other constructs a predictive latent under fixed constraints: LiLa-WAM~\citep{yang2026lilawamlightweightlatentreasoning} compresses current features with learned queries but decodes its predicted future into frozen DINOv3 features, while StageWAM~\citep{liu2026stagewamjointembeddingstageprediction} freezes V-JEPA2 on both sides and then freezes its predictor before policy learning. In both cases, the predictive target space does not co-evolve with the policy that ultimately consumes it.

To instantiate this perspective, we introduce \textbf{\emph{Copper-Policy}}, which treats predictive representation learning as part of the policy itself. Rather than requiring the model to reconstruct detailed futures, Copper-Policy distills task-relevant scene evolution into a compact latent representation and uses it to guide action generation. The representation is therefore encouraged to capture \emph{what is changing and where the task is heading}, while execution-specific spatial detail can remain available to the Action Expert. In this way, joint-embedding prediction serves as a mechanism for organizing representations for control, rather than as an additional generative objective.

\paragraph{Contributions.}
\begin{itemize}[leftmargin=1.3em,itemsep=1pt,topsep=2pt]

\item \textbf{Policy-learned predictive target space.}
We formulate latent world modeling such that the predictive target space is learned jointly with the policy, rather than prescribed by a pretrained encoder or a staged representation. This lets predictive learning and action supervision jointly organize the representation used for control.

\item \textbf{Asymmetric prediction--control interface.}
We design the Future Expert to operate only on the compact predictive representation, while the Action Expert additionally accesses current-frame features. This allows the predictive target space to focus on task-relevant evolution without being forced to retain all execution-level spatial detail.

\item \textbf{Empirical and representation-level evidence.}
Across simulation and real-robot evaluations, Copper-Policy combines strong control performance with efficient training. Ablations and representation diagnostics link robustness to common closed-loop shifts to the learned World representation and show the complementary role of current-frame features.

\end{itemize}

%% file: sec_ARXIV/2_related_works.tex
\section{Related Work}
\label{sec:related}

\begin{figure*}[!t]
    \centering
    \includegraphics[width=\linewidth]{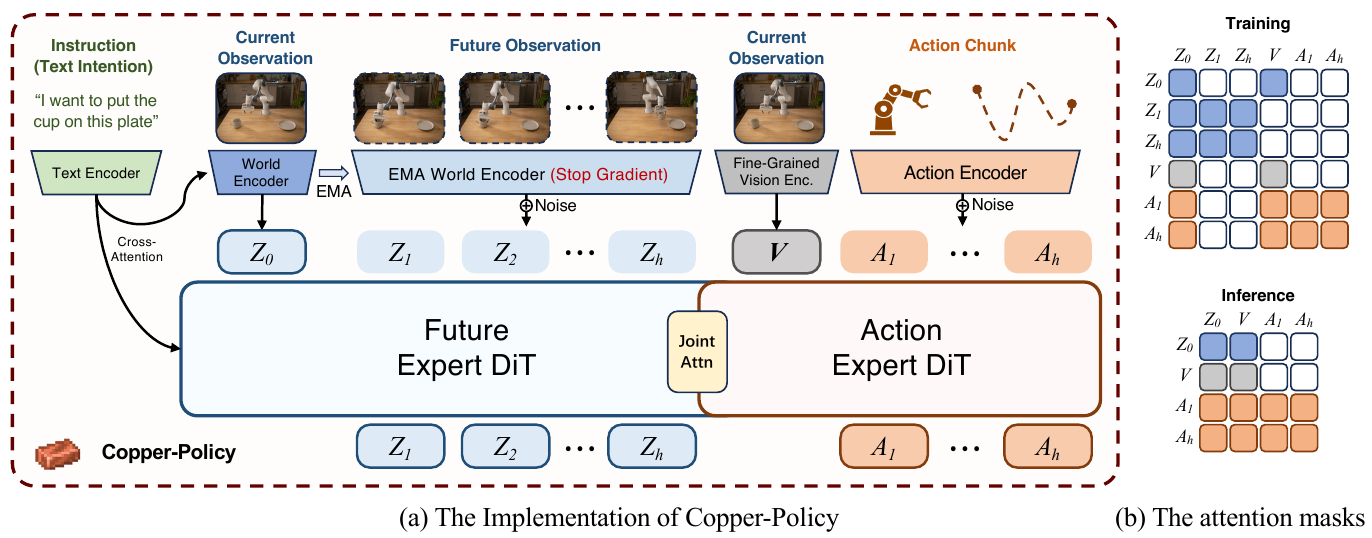}
    \caption{\textbf{Copper-Policy overview.} An online compactor produces the World representation $z_t$ (blue), while a spatial encoder and grid sampler produce current-frame tokens $v_t$ (gray). An EMA compactor supplies stop-gradient targets $\tilde z_{t+1:t+H_f}$. The Future Expert predicts these targets, and the Action Expert decodes the action chunk (orange) with access to the robot state $q_t$.}
    \label{fig:method}
\end{figure*}

A broad line of work augments robot policy learning with signals derived from future observations or prediction, spanning world action models, future-conditioned policies, and joint-embedding prediction. A key distinction is the target space, which determines what information the representation retains for policy. Our comparison concerns representations that the policy reads for action decoding. In Copper-Policy, the predictive target space is the same compact World representation read by the Action Expert.

\paragraph{Fixed target spaces preserve information beyond what control requires.}
Fast-WAM~\citep{yuan2026fastwam}, LingBot-VA~\citep{li2026causal}, and Cosmos-Policy~\citep{kim2026cosmos} predict future visual targets in pretrained video-generation latents, whereas earlier video-prediction policies generate future images or videos directly~\citep{NEURIPS2023_1d5b9233,ICLR2024_b2d4051f,cheang2024gr}. Both substrates preserve information needed for visual synthesis and may retain appearance not relevant to action prediction. Other WAM targets include future RGB--D videos~\citep{guo2026unified}, evolving geometric relations~\citep{zhang2026learning}, and RGB--depth--flow futures~\citep{zhao2026rynnworld}; outside WAMs, the same pattern appears with targets that are not futures at all~\citep{spatialforcing2025}.

\paragraph{Control-oriented targets compress future information, but are typically pretrained or staged.}
A second line uses future observations to learn control-oriented policy states. FLARE~\citep{pmlr-v305-zheng25a} carries future tokens through action denoising and aligns them with the future-observation embedding of an action-aware vision-language encoder, pretrained in an earlier stage and kept as a stop-gradient EMA target encoder during policy training. WoG~\citep{su2026world} first learns a compact future condition through action prediction, then freezes it while training the current-context pathway to predict that condition. Being-H0.7~\citep{beingbeyond2026beingh07} instead jointly aligns current latent queries with a future-informed posterior inside the action model. Together, these methods show that privileged future information can shape compact, action-relevant policy states without test-time rollout. Our predictive target space is learned jointly with the policy. Current and future representations inhabit the same compact space, which action decoding also uses and shapes alongside prediction. Robot-relevant visual representations can also be pretrained on human and other Internet videos through time-contrastive~\citep{pmlr-v205-nair23a}, value-based~\citep{ma2023vip}, masked~\citep{pmlr-v205-radosavovic23a}, or language-driven~\citep{Karamcheti-RSS-23} objectives and then frozen. A separate line learns latent \emph{actions} from unlabeled video~\citep{ICLR2025_45d74e19,pmlr-v235-bruce24a}, encoding transitions rather than a compact observation representation.

\paragraph{JEPA predicts in representation space, but prior robot policies do not jointly shape that space through control.}
Joint-embedding predictive architectures (JEPAs) predict targets in representation space rather than reconstructing pixels. I-JEPA~\citep{Assran_2023_CVPR} and V-JEPA~\citep{bardes2024revisiting} use momentum-updated target encoders to define masked representation targets in images and videos, learning visual representations without an action objective and with deterministic predictors. VLA-JEPA~\citep{sun2026vla} first pretrains a student on current observations to predict latent representations produced by a target encoder from future frames, then fine-tunes an action head. Copper-Policy instead learns the predictive target space jointly with action decoding and predicts in that same space by flow matching. Latent world models~\citep{ha2018world,hafner2019learning,Hafner2025} put related predictive structures to a different use. DINO-WM~\citep{pmlr-v267-zhou25t} and LeWM~\citep{maes2026leworldmodel} roll out frozen patch features and a learned latent, conditioned on candidate actions so that a planner can search over them at test time.

Copper-Policy builds on Fast-WAM's two-stream design and its finding that future prediction need not be used at deployment. The key difference is that the predictive target space is learned jointly with action decoding, so prediction shapes the same compact representation used for control.

%% file: sec_ARXIV/4_method.tex
\section{Method}
\label{sec:method}

\subsection{Overview}
\label{sec:overview}

\Cref{fig:method} gives the architecture. At time $t$ the policy observes $o_t$, a set of camera views, together with robot state $q_t$ and task instruction $s$, and produces an action chunk of horizon $H_a$,
\begin{equation}
    a^+ := a_{t:t+H_a-1} \sim p(a^+ \mid o_t, q_t, s).
\end{equation}
WAMs augment this with prediction targets over a future horizon $H_f$. Copper-Policy uses temporal joint-embedding prediction to learn a compact space conditioned on task intention and constrained by action decoding, implemented by an online--EMA compactor pair and two expert streams coupled through a structured attention mask. No future is predicted at test time.

\input{sec_ARXIV/4_setting_tables}

\subsection{Shared Representation with Online and EMA Compactors}
\label{sec:compactors}

Let $f$ map an observation into a source space, either the features of a frozen pretrained encoder or the pixels themselves, and let $e_s = E_{\text{text}}(s)$ be a task-intention embedding from a text encoder, shared across the Future and Action Experts. The online compactor $C_\phi$ reads that source space under cross-attention to $e_s$ and writes a compact target space:
\begin{equation}
    z_t = C_\phi\!\big(f(o_t), e_s\big) \in \mathbb{R}^{K\times d},
\end{equation}
that is, $K$ compact tokens of width $d$. The input layout and the compactor's spatial mask determine $K$, yielding 28 tokens for LIBERO's multi-view layout and 36 for the T-shaped RoboTwin and real-robot layout. Band-limited queries keep a coarse correspondence between compact tokens and image regions. Here $f$ is a frozen V-JEPA-family encoder~\citep{bardes2024revisiting}, while $C_\phi$ is trainable. Together, $f$ and $C_\phi$ form the World Encoder of \cref{fig:method}.

We regard $o_t$ and $o_{t+\tau}$ from the same instructed trajectory as current and future views of one scene. Because the compactor reads $e_s$ directly, the compact space can retain what the instructed task depends on while discarding the rest. We call $z_t$ the \emph{World representation}, a compact, task-conditioned behavioral prior assigned to the current view in a shared temporal embedding space. It serves as the \emph{World input} to both expert streams in place of dense visual tokens.

Targets come from an EMA copy $C_{\phi'}$ of the same compactor, under stop-gradient:
\begin{equation}
    \tilde z_{t+\tau} = \mathrm{sg}\!\left(C_{\phi'}\!\big(f(o_{t+\tau}), e_s\big)\right), \qquad \tau = 1,\dots,H_f.
\end{equation}
Since a learned target space admits collapse, we follow BYOL~\citep{grill2020bootstrap} and subsequent joint-embedding practice~\citep{Assran_2023_CVPR}. Gradients are stopped through the target branch, and $C_{\phi'}$ tracks $C_\phi$ by EMA, so the target space co-evolves with the online representation. The frozen $f$ is shared rather than copied. The future horizon matches the action horizon, $H_f=H_a$, and we construct $N_f$ target nodes from this horizon. We write $\tilde z$ for the resulting target sequence and $z^{(u)}$ for its noised counterpart below.

\subsection{Temporal Joint-Embedding Prediction}
\label{sec:future_expert}

The Future Expert implements temporal joint-embedding prediction by predicting the compact future-view targets $\tilde z$ from $z_t$ via conditional flow matching~\citep{lipman2023flow}, a formulation also used by diffusion and flow policies~\citep{Chi-RSS-23,black2024pi_0}. The prediction is conditioned on task intention $e_s$. We parameterize the flow from the clean target at $u=0$ to pure noise at $u=1$. For noise $\epsilon\sim\mathcal N(0,I)$, we interpolate
\begin{equation}
    z^{(u)} = (1-u)\,\tilde z + u\,\epsilon,
\end{equation}
and train a velocity field $\pi_\theta(z^{(u)}, u \mid z_t, e_s)$, the Future Expert itself, to match $\epsilon - \tilde z$, the derivative of the interpolation path with respect to $u$. The flow-matching Future Expert is one of two streams in a joint self-attention transformer based on the mixture-of-transformers pattern~\citep{liang2025mixtureoftransformers,yuan2026fastwam}. A structured mask lets it attend only to $z_t$ and other future tokens, not directly to the current-frame tokens $v_t$ or the robot state introduced next.

\subsection{Action Expert and Current-Frame Vision}
\label{sec:action_expert}

A separate pathway supplements $z_t$ with current-frame spatial detail. A learnable spatial encoder $g_\psi$ feeds a differentiable grid sampler $S_\xi$~\citep{feng2026see}, a visual bottleneck that compresses a feature map into a few tokens, and together they produce $v_t = S_\xi(g_\psi(o_t))$. The Action Expert, the second stream of the joint self-attention transformer, additionally receives the robot state $q_t$ as a single token projected alongside $e_s$, and decodes the action chunk with the same flow-matching formulation. For $a^{(u)} = (1-u)\,a^+ + u\,\epsilon_a$ it predicts the velocity
\begin{equation}
    \hat v^{a} = \pi^{a}_\theta\!\left(a^{(u)}, u \mid z_t, v_t, q_t, e_s\right),
\end{equation}
trained to match $\epsilon_a - a^+$ and integrated at test time to give $\hat a^+$.
Under the same mask, action tokens attend to $z_t$, to $v_t$, and to other action tokens, but never to the predicted future tokens. Thus, temporal joint-embedding prediction affects control only through the representations it shapes during training. Because the compactor is shared, the action loss directly constrains the space itself, pressing the World representation to carry what the action decoder reads.

\input{sec_ARXIV/5_libero_plus_table}

\subsection{Asymmetric Interface}
\label{sec:asymmetry}

This asymmetric interface keeps the predictive target compact while giving control direct access to current-frame detail. The structured mask of \cref{sec:future_expert,sec:action_expert} gives the two objectives unequal readouts. Future queries attend to $z_t$ and other future tokens, whereas action queries attend to $z_t$, the current-frame tokens $v_t$, and the state $q_t$. Within the current-step prefix, $z_t$ and $v_t$ attend to each other, so detail reaches future queries only after passing through the $K$ compact tokens, while action decoding reads $v_t$ directly. Consequently, joint-embedding prediction explains the future from $z_t$~\citep{grill2020bootstrap,Assran_2023_CVPR}, and its targets contain only compactor outputs. Action decoding can therefore use instantaneous detail without requiring $z_t$ to duplicate it, while the action loss still shapes $z_t$ toward information useful for control.

We test two predictions in \cref{sec:rep_analysis}. Joint-embedding prediction should reduce the overlap between perturbation-induced and task-driven temporal variation in $z_t$. The detail needed for precise action should live in $v_t$ rather than in $z_t$, even though the action loss also presses on $z_t$.

\subsection{Compact Token Interface}
\label{sec:token_interface}

Each future target step adds a block of target tokens, so the target space sets the sequence length the experts attend over at every training step. In the real-robot setting Copper-Policy processes 244 tokens per sample against 392 for Fast-WAM, and the Future Expert attends over 180 of them against 360. Self-attention is quadratic in that length, so the size of the target space bounds training throughput as the prediction horizon grows.

\subsection{Joint Objective and Inference}
\label{sec:objective}

The temporal joint-embedding loss is
\begin{equation}
    \mathcal{L}_{\text{pred}} = \mathbb{E}\left\| \pi_\theta\!\left(z^{(u)}, u \mid z_t, e_s\right) - \left(\epsilon-\tilde z\right)\right\|_2^2.
\end{equation}
and the action loss regresses the same kind of velocity. We optimize the action and prediction objectives jointly; $C_{\phi'}$ never receives gradients.

\paragraph{Inference.}
\label{sec:inference}
At test time, the model receives only the current observation $o_t$, the robot state $q_t$, and the instruction $s$. The online compactor and grid sampler produce $z_t$ and $v_t$; the Future Expert stream runs once, as a context encoder, to populate its per-layer keys and values. The Action Expert then decodes $\hat a^+$ over 10 flow steps, attending jointly over that cache and its own tokens. No future tokens are instantiated.

%% file: sec_ARXIV/4_setting_tables.tex
\begin{table}[!t]
\centering
\scriptsize
\setlength{\tabcolsep}{8pt}
\renewcommand{\arraystretch}{0.90}
\begin{tabular}{lccc}
\toprule
 & LIBERO & RoboTwin & Real \\
\midrule
Camera views & 2 & 3 & 3 \\
Action space & delta EEF & abs.\ qpos & abs.\ qpos \\
Demos / task & 50 & 50 & 300 \\
Rollouts / task & 50 & 100 & 50 \\
\bottomrule
\end{tabular}
\caption{Per-benchmark data and rollout settings. The action space is either delta end-effector (EEF) pose or absolute joint positions (qpos).}
\label{tab:configs}
\end{table}

\begin{table}[!t]
\centering
\scriptsize
\setlength{\tabcolsep}{4pt}
\renewcommand{\arraystretch}{0.90}
\begin{tabular}{lccc}
\toprule
Configuration & $z_t$ & JE Pred. & $v_t$ \\
\midrule
Copper-Policy & $\checkmark$ & $\checkmark$ & $\checkmark$ \\
\hspace{0.5em}--~w/o Joint-Embedding Prediction & $\checkmark$ & $\times$ & $\checkmark$ \\
\hspace{0.5em}--~w/o Current-Frame Visual Features & $\checkmark$ & $\checkmark$ & $\times$ \\
\hspace{0.5em}--~w/o World Input \& Future Expert & $\times$ & $\times$ & $\checkmark$ \\
\hspace{0.5em}--~w/o JE Pred.\ \& Visual Features & $\checkmark$ & $\times$ & $\times$ \\
\bottomrule
\end{tabular}
\caption{Ablation configurations. \emph{w/o World Input \& Future Expert} gives the Action Expert only current-frame features and language.}
\label{tab:ablation_config}
\end{table}

%% file: sec_ARXIV/5_libero_plus_table.tex
\begin{table*}[!t]
\centering
\scriptsize
\setlength{\tabcolsep}{2pt}
\renewcommand{\arraystretch}{0.90}
\resizebox{\textwidth}{!}{%
\begin{tabular}{p{110pt}|cc|c@{\hspace{5.5pt}}c@{\hspace{5.5pt}}c@{\hspace{5.5pt}}c@{\hspace{5.5pt}}c|ccccc@{\hspace{2.5pt}}c@{\hspace{2.5pt}}c@{\hspace{2.5pt}}c@{}}
\toprule
\multirow{2}{*}{\raisebox{-2.5pt}{\makebox[110pt][l]{Method}}} & \multirow{2}{*}{\raisebox{-8pt}{\shortstack[c]{Embodied\\Pretrain}}} & \multirow{2}{*}{\raisebox{-2.5pt}{WAM}} & \multicolumn{5}{c|}{LIBERO} & \multicolumn{8}{c}{LIBERO-Plus} \\[-1pt]
\cmidrule(lr){4-8}\cmidrule(l){9-16}
& & & Long & Goal & Object & Spatial & Avg. & Camera & Robot & Language & Light & Background & Noise & Layout & Total \\[-1pt]
\midrule
\textbf{Copper-Policy (ours)} & $\times$ & $\checkmark$\supmark{\ddagger} & 95.2 & 95.6 & 99.4 & \textbf{98.8} & 97.25 & \textbf{78.67} & 73.74 & 67.34 & 95.27 & 84.20 & \textbf{94.44} & 76.52 & 80.85 \\
$\pi_{0.5}$~\citep{intelligence2025pi_} & $\checkmark$ & $\times$ & 92.4 & 98.0 & 98.2 & \textbf{98.8} & 96.9 & 75.4 & \underline{77.5} & 85.6 & \textbf{96.9} & \underline{94.6} & 89.7 & \textbf{85.7} & \textbf{85.7} \\
Cosmos-Policy~\citep{kim2026cosmos} & $\times$ & $\checkmark$ & \textbf{97.6} & \underline{98.2} & \textbf{100.0} & 98.1 & \underline{98.5} & \underline{75.8} & 63.3 & 81.7 & \underline{96.5} & 88.9 & \underline{92.7} & 82.2 & \underline{82.2} \\
ABot-M0~\citep{yang2026abot} & $\checkmark$ & $\times$ & \underline{96.6} & \textbf{99.0} & \underline{99.8} & \textbf{98.8} & \textbf{98.6} & 60.4 & 67.9 & \underline{86.4} & 96.2 & 91.6 & 86.4 & 82.6 & 80.5 \\
VLA-JEPA~\citep{sun2026vla} & $\checkmark$ & $\checkmark$\supmark{\ddagger} & 95.8 & 97.2 & 99.6 & 96.2 & 97.2 & 64.2 & 67.7 & \textbf{88.1} & 91.8 & 93.4 & 65.8 & \underline{83.9} & 77.9 \\
X-VLA~\citep{zheng2026x} & $\checkmark$ & $\times$ & \textbf{97.6} & 97.8 & 98.6 & \underline{98.2} & 98.1 & 23.4 & \textbf{89.7} & 75.7 & 88.2 & \textbf{96.0} & 62.7 & 71.8 & 71.4 \\
Fast-WAM~\citep{yuan2026fastwam} & $\times$ & $\checkmark$\supmark{\ddagger} & 95.2 & 97.0 & \textbf{100.0} & \underline{98.2} & 97.6 & 16.4 & 44.5 & 68.9 & 78.2 & 53.7 & 37.7 & 60.7 & 51.5 \\
\midrule
\multicolumn{16}{l}{\raisebox{0.5pt}{\textit{Copper-Policy Variants}}} \\[-1pt]
\midrule
\multicolumn{3}{l|}{Copper-Policy} & \textbf{95.2} & \textbf{95.6} & \underline{99.4} & \textbf{98.8} & \textbf{97.25} & \textbf{78.67} & \underline{73.74} & \textbf{67.34} & \textbf{95.27} & 84.20 & \textbf{94.44} & \textbf{76.52} & \textbf{80.85} \\
\multicolumn{3}{l|}{\hspace{0.5em}--~w/o Joint-Embedding Prediction} & \underline{91.2} & 95.0 & \textbf{100.0} & \underline{97.4} & \underline{95.90} & 76.17 & \textbf{78.58} & \underline{53.94} & 93.52 & \textbf{90.33} & \underline{91.63} & \underline{76.26} & \underline{79.11} \\
\multicolumn{3}{l|}{\hspace{0.5em}--~w/o Current-Frame Visual Features} & 90.6 & \underline{95.4} & 99.0 & 96.8 & 95.45 & \underline{76.49} & 71.23 & 52.77 & 95.01 & 74.26 & 81.32 & 72.72 & 74.11 \\
\multicolumn{3}{l|}{\hspace{0.5em}--~w/o World Input \& Future Expert} & 90.8 & 92.4 & 98.8 & 97.2 & 94.80 & 66.79 & 67.23 & 52.18 & \underline{95.18} & \underline{88.20} & 87.07 & 74.49 & 74.56 \\
\multicolumn{3}{l|}{\hspace{0.5em}--~w/o Joint-Embedding Prediction \& Visual Features} & 66.8 & 92.6 & 89.4 & 91.2 & 85.00 & 63.73 & 62.26 & 30.97 & 86.51 & 52.32 & 55.53 & 65.51 & 58.81 \\
\bottomrule
\end{tabular}
}
\caption{LIBERO and LIBERO-Plus success rates (\%). WAM marks explicit future modeling, and $^{\ddagger}$ that no future is generated at deployment. Here and in later tables, bold and underlining mark the best and second best within a block, ties included.}
\label{tab:libero_plus_ablation}
\end{table*}

%% file: sec_ARXIV/5_experiments.tex
\section{Experiments}
\label{sec:experiments}

We evaluate Copper-Policy on LIBERO and LIBERO-Plus, RoboTwin, and three real-robot manipulation tasks, using the same ablation configurations throughout (Table~\ref{tab:ablation_config}).

\subsection{Experimental Settings}

\paragraph{LIBERO and LIBERO-Plus.}
Following the standard LIBERO protocol~\citep{NEURIPS2023_8c3c6668}, we evaluate all 40 tasks across the Spatial, Object, Goal, and Long suites, using 50 rollouts per task and the same closed-loop replanning protocol for every variant. For LIBERO-Plus~\citep{fei2025libero}, we report task-weighted success over 10{,}030 perturbed tasks. Comparison numbers are from the respective papers or from the evaluation of \citet{zhang2026world}.

\paragraph{RoboTwin.}
We evaluate each variant on all 50 official tasks, with 100 rollouts per task in each setting. \emph{Clean2Clean} trains and evaluates on the clean set, whereas \emph{Clean2Rand} evaluates clean-trained policies under joint randomization of layout, distractors, background, and lighting. Comparison numbers are from the benchmark leaderboard~\citep{chen2025robotwin} and the respective papers.

\paragraph{Real robot.}
We evaluate \emph{Fold the Towel} for long-horizon deformable-object manipulation, \emph{Put Banana} for pick-and-place under adversarial object placements, and \emph{Stack Three Bowls} for multi-stage rigid-object manipulation. The platform comprises two 6-degree-of-freedom arms with standard grippers and three RGB cameras; no depth input is used. Actions are represented as 14-D absolute joint positions (qpos). We collect 300 teleoperated demonstrations per task at 30\,Hz, totaling approximately 4.3 hours. An automatic scene generator produces 50 seeded initial configurations per task, on which each policy is evaluated using real-time chunking~\citep{black2025training}.

\subsection{Main Results}
\subsubsection{LIBERO and LIBERO-Plus}
\label{sec:libero_results}
\label{sec:libero}

Copper-Policy averages 97.25\% on standard LIBERO. Its 80.85\% LIBERO-Plus total exceeds several embodied-pretrained VLA baselines; only Cosmos-Policy (82.2\%), which fine-tunes Cosmos-Predict2 as its policy backbone, and large-scale embodied-pretrained $\pi_{0.5}$ (85.7\%) score higher. Cosmos-Policy's small overall lead is driven by Language, as Copper-Policy has the higher pooled success rate over the other six categories. It leads Camera and Noise, ranks behind only X-VLA and $\pi_{0.5}$ on Robot among comparison methods, and remains strong on Light.

\subsubsection{RoboTwin}
\label{sec:robotwin}

Among methods without embodied pretraining, Copper-Policy achieves the highest average success rate at 41.91\%, ahead of EventVLA at 40.65\% and Fast-WAM at 39.85\%. Its clean-set score is on par with $\pi_{0.5}$ (70.84\% versus 70.70\%). Under randomization, however, it remains substantially below $\pi_{0.5}$ (12.98\% versus 46.00\%).

\begin{table}[t]
\centering
\scriptsize
\setlength{\tabcolsep}{2pt}
\renewcommand{\arraystretch}{0.90}
\resizebox{\columnwidth}{!}{%
\begin{tabular}{p{80pt}|cc|ccc}
\toprule
\multirow{2}{*}{\raisebox{-2.5pt}{\makebox[80pt][c]{Method}}} & \multirow{2}{*}{\raisebox{-2.5pt}{\shortstack[c]{Embodied\\Pretrain}}} & \multirow{2}{*}{\raisebox{-2.5pt}{WAM}} & \multicolumn{3}{c}{RoboTwin} \\
\cmidrule(l){4-6}
& & & Clean & Random & Avg. \\
\midrule
\multicolumn{6}{l}{\textit{Comparison with methods using embodied pretraining}} \\
\midrule
$\pi_{0.5}$~\citep{intelligence2025pi_} & $\checkmark$ & $\times$ & 70.70 & \textbf{46.00} & \textbf{58.35} \\
Spatial Forcing~\citep{spatialforcing2025} & $\checkmark$ & $\times$ & \textbf{77.20} & 26.74 & \underline{51.97} \\
X-WAM~\citep{guo2026unified} & $\checkmark$ & $\checkmark$ & 70.00 & 25.80 & 47.90 \\
X-VLA~\citep{zheng2026x} & $\checkmark$ & $\times$ & 68.00 & 20.90 & 44.45 \\
ABot-M0~\citep{yang2026abot} & $\checkmark$ & $\times$ & 57.40 & \underline{30.36} & 43.88 \\
\textbf{Copper-Policy (ours)} & $\times$ & $\checkmark$\supmark{\ddagger} & \underline{70.84} & 12.98 & 41.91\supmark{\dagger} \\
Xiaomi Robotics-0~\citep{cai2026xiaomi} & $\checkmark$ & $\times$ & 62.90 & 18.20 & 40.55 \\
GalaxeaVLA~\citep{liu2026g05autoregressivestreamrobot} & $\checkmark$ & $\times$ & 62.70 & 12.72 & 37.71 \\
RDT-1B~\citep{ICLR2025_49f80e4d} & $\checkmark$ & $\times$ & 34.50 & 13.72 & 24.11 \\
\midrule
\multicolumn{6}{l}{\textit{Methods without embodied pretraining}} \\
\midrule
\textbf{Copper-Policy (ours)} & $\times$ & $\checkmark$\supmark{\ddagger} & \underline{70.84} & \underline{12.98} & \textbf{41.91} \\
EventVLA~\citep{yang2026eventvla} & $\times$ & $\times$ & 65.60 & \textbf{15.70} & \underline{40.65} \\
Fast-WAM~\citep{yuan2026fastwam} & $\times$ & $\checkmark$\supmark{\ddagger} & \textbf{77.80} & 1.90 & 39.85 \\
AHA-WAM~\citep{cai2026aha} & $\times$ & $\checkmark$ & 64.30 & 3.20 & 33.75 \\
DP3~\citep{Ze-RSS-24} & $\times$ & $\times$ & 55.24 & 4.96 & 30.10 \\
starVLA~\citep{ye2026starvla} & $\times$ & $\times$ & 46.52 & 3.16 & 24.84 \\
ACT~\citep{Zhao-RSS-23} & $\times$ & $\times$ & 29.74 & 1.74 & 15.74 \\
\midrule
\multicolumn{6}{l}{\textit{Copper-Policy Variants}} \\
\midrule
\multicolumn{3}{l|}{Copper-Policy} & \textbf{70.84} & \textbf{12.98} & \textbf{41.91} \\
\multicolumn{3}{l|}{\hspace{0.5em}--~w/o Joint-Embedding Prediction} & 48.16 & 9.38 & 28.77 \\
\multicolumn{3}{l|}{\hspace{0.5em}--~w/o Current-Frame Visual Features} & \underline{67.82} & \underline{10.22} & \underline{39.02} \\
\multicolumn{3}{l|}{\hspace{0.5em}--~w/o World Input \& Future Expert} & 55.26 & \underline{10.22} & 32.74 \\
\multicolumn{3}{l|}{\hspace{0.5em}--~w/o Joint-Embedding Prediction \& Visual Features} & 51.66 & 9.42 & 30.54 \\
\bottomrule
\end{tabular}
}
\caption{RoboTwin success rates (\%). In the upper block, $^{\dagger}$ marks the one entry trained without embodied pretraining; $^{\ddagger}$ marks methods that generate no future at deployment.}
\label{tab:robotwin_ablation}
\end{table}

\begin{figure*}[!t]
\centering
\makebox[\textwidth][c]{%
\begin{minipage}[t]{0.40\textwidth}
\vspace*{0pt}
\raggedleft
\setlength{\tabcolsep}{0pt}
\begin{tabular}{@{}l@{\hspace{3pt}}c@{}}
\rotatebox{90}{\tiny Towel} & \includegraphics[width=0.91\linewidth]{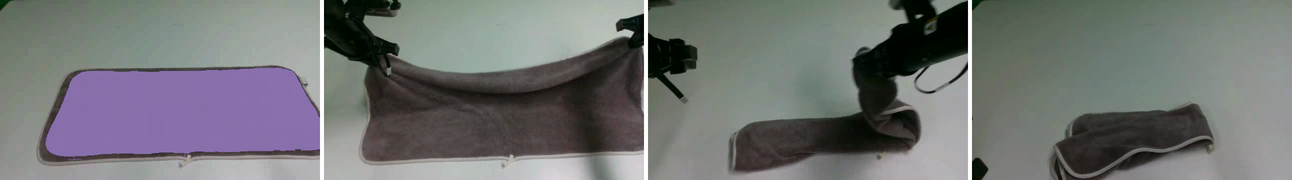} \\
\rotatebox{90}{\tiny Banana} & \includegraphics[width=0.91\linewidth]{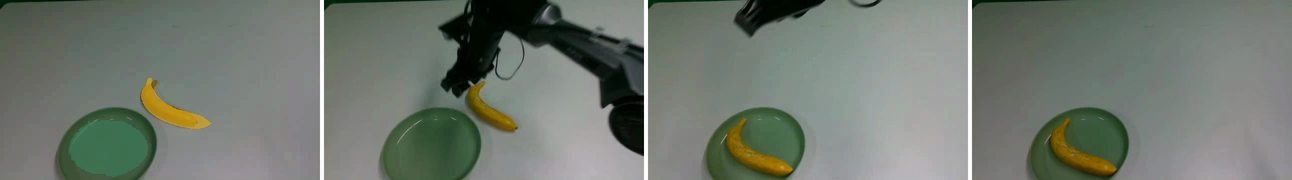} \\
\rotatebox{90}{\tiny Bowls} & \includegraphics[width=0.91\linewidth]{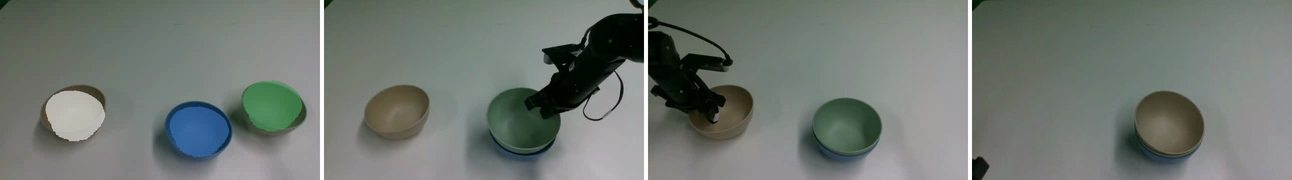} \\
\end{tabular}
\end{minipage}\hspace{20pt}%
\begin{minipage}[t]{0.52\textwidth}
\vspace*{0pt}
\raggedright
\footnotesize
\setlength{\tabcolsep}{5pt}
\renewcommand{\arraystretch}{0.90}
\begin{tabular}{l|ccc|c}
\toprule
\shortstack[l]{Configuration} & Towel & Banana & Bowls & Avg. \\
\midrule
$\pi_{0.5}$ & \underline{95} & \underline{94} & \underline{94} & \underline{94.3} \\
Fast-WAM & \textbf{98} & 60 & \underline{94} & 84.0 \\
\midrule
\textbf{Copper-Policy} & 93 & \textbf{96} & \textbf{100} & \textbf{96.3} \\
w/o World Input \& Future Expert & 84 & 82 & 88 & 84.7 \\
w/o Joint-Embedding Prediction & 4 & 26 & \textbf{100} & 43.3 \\
w/o Current-Frame Visual Features & 2 & 46 & 14 & 20.7 \\
w/o JE Pred.\ \& Visual Features & 0 & 4 & 0 & 1.3 \\
\bottomrule
\end{tabular}
\end{minipage}
}
\caption{\textbf{Real-robot evaluation.} The left panel shows four frames from one successful rollout per task, with target poses overlaid in the first column. The right panel reports scores that assign 1, 0.5, and 0 to full, half, and no success. $\pi_{0.5}$ uses embodied pretraining.}
\label{fig:realbot_tasks}
\end{figure*}

\subsubsection{Real-Robot Evaluation}
\label{sec:real_robot}

Each policy uses the protocol and initial-scene seeds described above. Copper-Policy obtains the highest average, 96.3 against 94.3 for $\pi_{0.5}$ and 84.0 for Fast-WAM, and is best on both rigid-object tasks (\cref{fig:realbot_tasks}). Removing the World input and Future Expert drops the average from 96.3 to 84.7. Fast-WAM performs best on towel folding at 98 but reaches 60 on banana placement, where it often does not attempt the grasp.

\subsection{Efficiency}
\label{sec:efficiency}

\input{sec_ARXIV/5_efficiency_table}

Predicting in a compact embedding space rather than generating video shortens the training sequence and permits smaller experts, putting the whole model at 2B trainable parameters. At a matched global batch of 128 on eight A100 GPUs, it uses less memory than $\pi_{0.5}$ and Fast-WAM and trains about $2\times$ and $6\times$ faster. The same configuration completes 30K steps in 9.67 hours on eight consumer RTX 5090 GPUs with plain data parallelism. End-to-end latency is about 85\,ms, against 75\,ms for $\pi_{0.5}$ and 80\,ms for Fast-WAM. This slightly higher latency reflects the frozen 300M V-JEPA encoder in the World Encoder, which runs at every step.

Within Copper-Policy, optimizer sharding lowers memory from 31.83 to 24.53\,GB while raising relative training time from 1.00$\times$ to 1.08$\times$; full sharding lowers memory to 21.24\,GB but raises time to 1.50$\times$ (\cref{tab:efficiency}). Copper-Policy fits on an RTX 5090 without sharding, whereas the two baselines use more memory despite full sharding. These values are operational cost references rather than a controlled systems benchmark because the methods differ in model size, sequence length, distributed strategy, and implementation.

\subsection{Controlled Comparisons}
\label{sec:controlled_comparisons}

\paragraph{The compact World representation supports robustness to common closed-loop shifts.}
Camera, Robot, Light, and Noise test changes that commonly arise during closed-loop execution. Copper-Policy leads on Camera and Noise, remains strong on Light, and ranks behind only X-VLA and $\pi_{0.5}$ on Robot. Removing the World input and Future Expert lowers overall success from 80.85\% to 74.56\%, with clear losses on Camera, Robot, and Noise. Retaining the World input without joint-embedding prediction recovers much of the overall score, but still trails the full model on Camera and Noise while scoring higher on Robot. These comparisons support the compact World representation's value across common closed-loop shifts, while showing that predictive shaping benefits the shift categories differently.

\paragraph{Language, Background, and Layout expose the generalization gap.}
Language, Background, and Layout test broader instruction and scene generalization, where several embodied-pretrained methods lead. Language is the clearest gap at 67.34\%. Copper-Policy uses frozen T5 without a pretrained VLM for language--image understanding or large-scale embodied pretraining, consistent with this result. More diverse training may help; we will explore the representations these categories need.

\paragraph{Cross-benchmark evidence for the representation design.}
The variant without joint-embedding prediction stays close to the full model on LIBERO-Plus but falls from 41.91 to 28.77 on RoboTwin and from 96.3 to 43.3 on the real robot. Together, these ablations and the four closed-loop shift categories support the compact World representation's role in retaining task-relevant information across visual and robot changes. The larger losses on RoboTwin and the real robot suggest that this role matters more in physical control. Removing current-frame features lowers the LIBERO, LIBERO-Plus, and RoboTwin aggregate scores by at most 6.74 points, yet drops the real-robot average to 20.7, consistent with the need for current-frame detail in contact-rich execution (\cref{sec:rep_analysis}).

\input{sec_ARXIV/5_representation_analysis}

%% file: sec_ARXIV/5_efficiency_table.tex
\begin{table*}[t]
\centering
\scriptsize
\setlength{\tabcolsep}{2.5pt}
\renewcommand{\arraystretch}{1.10}
\resizebox{\textwidth}{!}{%
\begin{tabular}{lrccccrcc}
\toprule
\raisebox{0.65ex}{Method} & \shortstack[c]{Trainable\\params.} & \raisebox{0.65ex}{Training hardware} & \raisebox{0.65ex}{Parallelism} & \shortstack[c]{Peak mem.\\/ GPU $\downarrow$} & \shortstack[c]{Relative\\training time $\downarrow$} & \shortstack[c]{Samples\\/ s $\uparrow$} & \shortstack[c]{Est. time\\(30K steps) $\downarrow$} & \shortstack[c]{RTX 5090\\inference $\downarrow$} \\
\midrule
Fast-WAM~\citep{yuan2026fastwam} & 6B & 8$\times$ A100 80GB & FSDP Fully Shard & 62.42\,GB & 6.13$\times$ & 14.36 & 74.3\,h & $\sim$80\,ms \\
$\pi_{0.5}$~\citep{intelligence2025pi_} & 3.3B & 8$\times$ A100 80GB & FSDP Fully Shard & 38.65\,GB & 1.96$\times$ & 44.91 & 23.7\,h & \textbf{$\sim$75\,ms} \\
\textbf{Copper-Policy (ours)} & \textbf{2B} & 8$\times$ A100 80GB & DDP & 31.83\,GB & \underline{1.00$\times$} & \underline{88.0} & \underline{12.1\,h} & $\sim$85\,ms \\
\midrule
\multicolumn{9}{l}{\raisebox{0.5pt}{\textit{Copper-Policy under alternative memory-sharding strategies (8$\times$ A100 80GB)}}} \\[-1pt]
\midrule
Copper-Policy & 2B & 8$\times$ A100 80GB & DDP & 31.83\,GB & \underline{1.00$\times$} & \underline{88.0} & \underline{12.1\,h} & $\sim$85\,ms \\
Copper-Policy & 2B & 8$\times$ A100 80GB & DDP + Optimizer Shard & \underline{24.53\,GB} & 1.08$\times$ & 81.8 & 13.0\,h & $\sim$85\,ms \\
Copper-Policy & 2B & 8$\times$ A100 80GB & FSDP Fully Shard & \textbf{21.24\,GB} & 1.50$\times$ & 58.56 & 18.2\,h & $\sim$85\,ms \\
\midrule
\textbf{Copper-Policy (ours)} & \textbf{2B} & \textbf{8$\times$ RTX 5090 32GB} & DDP & 31.83\,GB & \textbf{0.80$\times$} & \textbf{110.34} & \textbf{9.67\,h} & $\sim$85\,ms \\
\bottomrule
\end{tabular}
}
\caption{Training cost at global batch 128, normalized to Copper-Policy under DDP on 8$\times$ A100 (1.454\,s/step). Peak memory is reserved memory per GPU. FSDP Fully Shard covers ZeRO-3, FSDP, and FSDP2 full-shard; DDP + Optimizer Shard is ZeRO-1. The middle block changes only Copper-Policy's sharding strategy. Estimated time assumes 30K steps; inference is end-to-end.}
\label{tab:efficiency}
\end{table*}

%% file: sec_ARXIV/5_representation_analysis.tex
\subsection{Representation Analysis}
\label{sec:rep_analysis}

\subsubsection{Temporal Structure and Channel Complementarity}

Because the formulation is a claim about a representation, we inspect $z_t$ directly. We feed frames from recorded training episodes through each variant offline and measure how $z_t$ changes over time, treating the episode rather than the frame as the statistical unit so that long episodes do not dominate. The analysis covers 40 episodes per variant on LIBERO (ten tasks from each of the four suites), 18 on the real robot (six per task), and 20 on RoboTwin (five from each of four tasks), all drawn with one fixed seed.

\begin{figure*}[t]
\centering
\includegraphics[width=\textwidth]{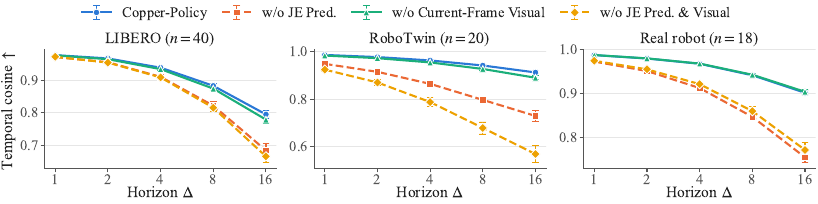}
\caption{Cosine similarity of unit-normalized $z_t$ across horizon $\Delta$ on LIBERO, RoboTwin, and the real robot, with 95\% confidence intervals over episodes.}
\label{fig:temporal_cosine}
\end{figure*}

\begin{table*}[t]
\centering
\scriptsize
\setlength{\tabcolsep}{6pt}
\renewcommand{\arraystretch}{1.1}
\begin{tabular}{l|cc|cc|cc}
\toprule
\multirow{2}{*}{\raisebox{-4.5pt}{Configuration}} & \multicolumn{2}{c|}{LIBERO ($n=40$)} & \multicolumn{2}{c|}{RoboTwin ($n=20$)} & \multicolumn{2}{c}{Real robot ($n=18$)} \\
\cmidrule(lr){2-3}\cmidrule(lr){4-5}\cmidrule(l){6-7}
& cosine $\uparrow$ & disp.\ $\downarrow$ & cosine $\uparrow$ & disp.\ $\downarrow$ & cosine $\uparrow$ & disp.\ $\downarrow$ \\
\midrule
\textbf{Copper-Policy} & \textbf{0.794} & \textbf{0.634} & \textbf{0.913} & \textbf{0.413} & \underline{0.902} & \underline{0.434} \\
\hspace{0.5em}--~w/o Joint-Embedding Prediction & 0.684 & 0.778 & 0.729 & 0.724 & 0.756 & 0.684 \\
\hspace{0.5em}--~w/o Current-Frame Visual Features & \underline{0.778} & \underline{0.660} & \underline{0.890} & \underline{0.463} & \textbf{0.904} & \textbf{0.429} \\
\hspace{0.5em}--~w/o Joint-Embedding Prediction \& Visual Features & 0.665 & 0.809 & 0.567 & 0.921 & 0.773 & 0.658 \\
\bottomrule
\end{tabular}
\caption{Temporal structure of the World representation, averaged over episodes. Cosine similarity and displacement are computed between unit-normalized representations 16 sampled frames apart; RoboTwin samples every fourth frame, so its separation spans more time and the three domains are not directly comparable. Bold and underlining mark the best and second best in each column.}
\label{tab:diagnostics}
\end{table*}

\paragraph{Joint-embedding prediction organizes the World representation in time.}
\Cref{tab:diagnostics} reports every variant at $\Delta=16$, and \cref{fig:temporal_cosine} traces cosine similarity across horizons. Removing joint-embedding prediction lowers cosine similarity in every domain, from $0.794$ to $0.684$ on LIBERO, from $0.902$ to $0.756$ on the real robot, and from $0.913$ to $0.729$ on RoboTwin, and raises displacement accordingly; without the joint-embedding objective the compactor still produces features, but they change faster over the episode. The paired comparison without $v_t$ keeps the ordering ($0.778$ versus $0.665$, $0.904$ versus $0.773$, and $0.890$ versus $0.567$). Across the complete attention set, joint-embedding prediction keeps $z_t$ focused on manipulated objects and other task-relevant regions as an episode proceeds, whereas without the objective the maps more often retain maxima on background surfaces and lack a stable task-relevant centre.

\paragraph{The diagnostics accompany a behavioral reversal.}
\emph{w/o Joint-Embedding Prediction} is worse than \emph{w/o World Input \& Future Expert} on RoboTwin ($28.77\%$ versus $32.74\%$) and on the real robot ($43.3$ versus $84.7$), even though it has strictly more machinery. An unshaped World pathway supplies the Action Expert with a signal that drifts quickly and lacks a stable task-relevant focus, which is worse for control than supplying no such signal at all. These measurements describe temporal smoothness of the representation; they are correlates of the mechanism, not proof of it.

\paragraph{The two pathways differ in temporal stability as well as spatial focus.}
Removing the current-frame pathway perturbs $z_t$ far less than removing joint-embedding prediction does: cosine similarity moves from $0.794$ to $0.778$ on LIBERO and from $0.913$ to $0.890$ on RoboTwin, and on the real robot the two are indistinguishable ($0.902$ versus $0.904$). Removing both joint-embedding prediction and the current-frame pathway is the largest drop on RoboTwin, though on the real robot it is no lower than removing joint-embedding prediction alone. Within the full model, $z_t$ is also more stable than the pooled current-frame tokens $v_t$ at the same separation, on LIBERO ($0.794$ versus $0.672$), on the real robot ($0.902$ versus $0.827$), and most clearly on RoboTwin ($0.913$ versus $0.245$). The current-frame tokens therefore change faster over an episode than the World representation does.

\subsubsection{Perturbation Geometry and Channel Probes}

We draw 25 offline episodes from each of four RoboTwin tasks (\emph{hanging mug}, \emph{place empty cup}, \emph{scan object}, and \emph{stack three blocks}), take 16 uniformly spaced valid frames per episode for 1{,}600 clean states, and pair every state with five deterministic image perturbations, namely camera translation, illumination change, global colour cast, Gaussian sensor noise, and blur, for 8{,}000 clean/perturbed pairs. All models run frozen in inference mode. Every split is by episode rather than by frame, and we report the mean and standard deviation over five deterministic episode-level splits. For a given rank $k$, we take $U_N$ and $U_T$, the leading $k$ directions of perturbation-induced and adjacent-frame temporal change, fit on training episodes alone, and report the overlap $\lVert U_N^\top U_T\rVert_F^2/k$ on held-out episodes, together with the energy of held-out perturbation change that falls in the temporal subspace. A random subspace of the same rank gives the baseline. Action probes use the same episode-level separation.

\begin{figure*}[t]
\centering
\includegraphics[width=\linewidth]{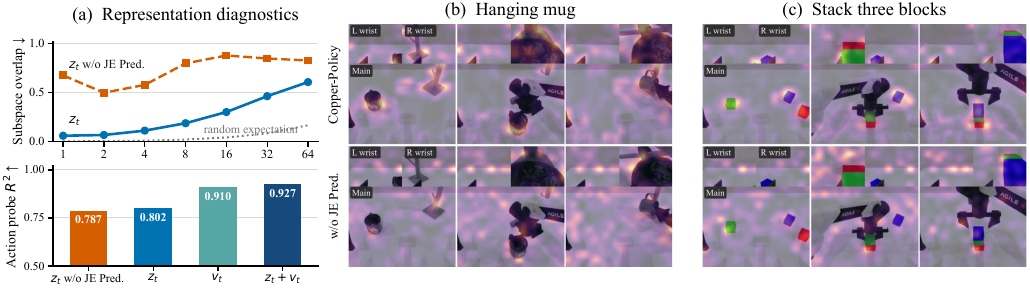}
\caption{\textbf{Representation analysis.} (a) Perturbation--temporal subspace overlap across rank (top; dotted line: random expectation) and linear action-probe $R^2$ (bottom) on 100 RoboTwin episodes. (b--c) $z_t$ attention across two RoboTwin episodes, comparing Copper-Policy (upper row) with \emph{w/o Joint-Embedding Prediction} (lower row).}
\label{fig:rep_real}
\end{figure*}

\begin{table*}[t]
\centering
\scriptsize
\setlength{\tabcolsep}{6pt}
\renewcommand{\arraystretch}{1.1}
\begin{tabular}{l|cc|cc|c}
\toprule
& \multicolumn{2}{c|}{Geometry at rank 8} & \multicolumn{2}{c|}{Paired perturbation} & Action probe \\
\cmidrule(lr){2-3}\cmidrule(lr){4-5}\cmidrule(l){6-6}
Representation & overlap $\downarrow$ & pert.\ in temp.\ $\downarrow$ & norm.\ $L_2$ $\downarrow$ & cosine $\uparrow$ & $R^2$ $\uparrow$ \\
\midrule
\textbf{Copper $z_t$} & \textbf{0.187} & \textbf{0.140} & \textbf{0.134} & \textbf{0.990} & 0.802 \\
\hspace{0.5em}--~w/o Joint-Embedding Prediction & 0.798 & 0.807 & 0.344 & 0.932 & 0.787 \\
Current-frame tokens $v_t$ & -- & -- & \underline{0.171} & \underline{0.969} & \underline{0.910} \\
Dense V-JEPA features & -- & -- & 0.428 & 0.904 & 0.831 \\
Token-matched V-JEPA features & -- & -- & 0.462 & 0.891 & 0.803 \\
\midrule
$z_t$ and $v_t$ together & -- & -- & -- & -- & \textbf{0.927} \\
\bottomrule
\end{tabular}
\caption{Representation diagnostics on 100 RoboTwin episodes. Geometry columns use train-only subspaces evaluated on held-out episodes; the random-subspace overlap is $0.021$ and random energy is $0.021$--$0.023$. Paired columns average 8{,}000 clean/perturbed pairs. Standard deviations over five episode-level splits are at most $0.006$ for geometry and $0.021$ for $R^2$.}
\label{tab:geometry}
\end{table*}

The first prediction holds. Joint-embedding prediction reorganizes rather than removes nuisance information. Perturbation identity remains linearly decodable from $z_t$ ($0.998$ balanced accuracy), yet the perturbation-change and temporal-change subspaces overlap far less with the objective across ranks (\cref{fig:rep_real}a, top). For the leading-$k$ perturbation-induced ($U_N$) and temporal-change ($U_T$) subspaces, overlap is $\lVert U_N^\top U_T\rVert_F^2/k$. At rank 8, it is $0.187$ versus $0.798$ without the objective and $0.021$ at random, with the same ordering on all four tasks. Across paired clean and perturbed views, the shaped $z_t$ also changes less in magnitude ($0.134$ versus $0.344$ mean normalized $L_2$ distance), while its attention remains on the manipulated object (\cref{fig:rep_real}b--c). Perturbations thus act along directions less aligned with task evolution, the representation-level counterpart of the Noise and Camera gains above.

The second geometry measure tells the same story. At rank 8, only $0.140$ of held-out perturbation-change energy lies in the temporal subspace of the learned $z_t$, compared with $0.807$ without joint-embedding prediction (\cref{tab:geometry}). This agrees with the low subspace overlap even though a probe can still identify the perturbation. The paired-view comparison adds a complementary measure: $z_t$ has a cosine similarity of $0.990$ between clean and perturbed images, compared with $0.932$ without the objective. Together, these measures show that the objective changes both the direction and magnitude of the representation's response to visual shifts.

The second prediction concerns the channels' division of information. On held-out clean episodes, a linear action probe yields $R^2$ of $0.802$ from $z_t$, $0.910$ from $v_t$, $0.927$ from both, and $0.787$ from $z_t$ without joint-embedding prediction (\cref{fig:rep_real}a, bottom). Instantaneous action detail is more directly accessible from $v_t$, and real-robot rollouts show why combining the channels matters. Without $v_t$, $z_t$ is the only visual input to both experts and remains as persistent as in the full model ($0.904$ versus $0.902$), yet the average task score falls from 96.3 to 20.7. Under our compact budget, the temporally shaped World channel cannot replace the current-frame pathway for precise control.

%% file: sec_ARXIV/6_conclusion.tex
\section{Conclusion}
\label{sec:conclusion}

Copper-Policy learns a compact World representation jointly with its policy through joint-embedding prediction and action decoding. Its asymmetric interface preserves current-frame detail for precise control. Without embodied pretraining, it performs strongly across simulation and real-robot tasks, substantially improving on the two-stream Fast-WAM in LIBERO-Plus and real-robot scores while training faster with fewer parameters. Diagnostics show reduced overlap between perturbation-induced and task-driven variation in $z_t$ and sustained attention to manipulated objects. Together with Camera, Robot, Light, and Noise results, this supports the representation's value under common closed-loop shifts.

\paragraph{Limitations.}
Strong Camera, Light, and Noise results despite the larger Language gap suggest the World representation is not narrowly tied to visual appearance. Instruction grounding remains limited, while Background and Layout reveal further scene-generalization gaps. Future work will learn representations that better capture task intent and scene structure across these settings.

%% file: sec_ARXIV/X_suppl.tex
\section{Token Composition}
\label{sec:supp_token_composition}

\subsection{Transformer Sequence Composition}
\label{sec:supp_tokens}

\begin{table}[H]
\centering
\small
\setlength{\tabcolsep}{12pt}
\renewcommand{\arraystretch}{1.1}
\begin{tabular}{l|ccc|c}
\toprule
Method & Current vis./world & Future & Action & Total \\
\midrule
$\pi_{0.5}$~\citep{intelligence2025pi_} & 768 & 0 & 32 & 800 \\
Fast-WAM~\citep{yuan2026fastwam} & 120 & 240 & 32 & 392 \\
\textbf{Copper-Policy} & \textbf{68} & \textbf{144} & 32 & \textbf{244} \\
\bottomrule
\end{tabular}
\caption{Transformer sequence composition per sample. For RoboTwin and real-robot runs, $|v_t|=32$ and the T-shaped visual layout yields $K=36$, giving $68$ current tokens; $N_f=4$ future target nodes give $144$. The multi-view LIBERO layout yields $K=28$. Of the current tokens the Future Expert sees only the $K$ World tokens, since the mask of \cref{sec:future_expert} excludes $v_t$. Each method uses 32 action tokens.}
\label{tab:tokens}
\end{table}

\section{Representation Analysis}
\label{sec:supp_diagnostics}

\subsection{Representation Diagnostics Protocols}
\label{sec:supp_diagnostic_protocols}

All representation diagnostics are read-only evaluations of frozen checkpoints. We run the model in evaluation mode with gradients disabled, use the same synchronized observations and language conditioning as in policy inference, and treat an episode—not an individual frame—as the statistical unit. For the World representation, the exported token and channel dimensions are flattened before temporal metrics are computed. For the current-frame pathway, the exported representation is the mean of its spatial tokens. Episode-level averaging gives each episode equal weight.

\paragraph{Temporal stability.}
For each episode we export the online World representation $z_t$ and, for the full model, the mean-pooled current-frame representation $v_t$. At horizons $\Delta\in\{1,2,4,8,16\}$ we compute cosine similarity and Euclidean displacement between unit-normalized representations,
\[
    c_\Delta=\frac{1}{T-\Delta}\sum_t \hat r_t^\top\hat r_{t+\Delta},\qquad
    d_\Delta=\frac{1}{T-\Delta}\sum_t\left\|\hat r_{t+\Delta}-\hat r_t\right\|_2.
\]
We average the resulting episode-level values and report 95\% confidence intervals across episodes. The main temporal comparison uses 40 LIBERO episodes, 20 RoboTwin episodes, and 18 real-robot episodes. RoboTwin is subsampled every fourth frame, so the same index separation spans a longer physical interval there; its absolute values are therefore not compared numerically with LIBERO or the real robot.

\paragraph{Perturbation geometry and information probes.}
The larger perturbation analysis uses 100 RoboTwin episodes: 25 episodes from each of four tasks, with 16 uniformly spaced valid frames per episode. Each clean frame is paired with five deterministic visual perturbations—camera translation, illumination change, global color cast, Gaussian sensor noise, and blur—yielding 1,600 clean states and 8,000 paired examples. Splits are made by episode, never by frame, and all reported numbers are means over five deterministic episode-level splits. For geometry at rank $k=8$, we fit the leading perturbation-change subspace $U_N$ and adjacent-frame temporal-change subspace $U_T$ on training episodes and evaluate
\[
    \frac{\left\|U_N^\top U_T\right\|_F^2}{k}
    \quad\text{and}\quad
    \frac{\|U_TU_T^\top\delta_N\|_2^2}{\|\delta_N\|_2^2}
\]
on held-out episodes. A random subspace of the same rank is the reference. Linear nuisance and action probes are fit on the training episodes only; the former predicts perturbation identity, while the latter predicts the action chunk. For $v_t$, we mean-pool current-frame tokens before probing. We also report the paired clean/perturbed $L_2$ distance and cosine similarity.

\paragraph{Qualitative protocol.}
For the project-page $z_t$ visualizations, the full model and \emph{w/o Joint-Embedding Prediction} receive identical frames, task instructions, camera ordering, and rendering schedules. We render six evenly spaced steps per selected episode and preserve the native multi-view layout.

\subsection{Attention-Map Generation}
\label{sec:supp_attention_generation}

For the $z_t$ attention shown in the main text and on the project page, we capture the attention probabilities in the final task-conditioned compactor read block. If $A_{hqp}$ denotes the softmax probability from compact query $q$ and head $h$ to input patch $p$, the scalar patch score is
\[
    m^{Z}_p=\frac{1}{HQ}\sum_{h=1}^{H}\sum_{q=1}^{Q}A_{hqp}.
\]
The patch sequence is then split according to the original camera grids, $m^Z_p\mapsto m^{Z,(m)}\in\mathbb{R}^{h_m\times w_m}$. The resulting $z_t$ attention is the final-compactor read attention averaged over heads and compact queries.

For the current-frame representation map, we compute a post-hoc source-patch response from the tokens produced by the coordinate-based bilinear sampler. Let $q_\ell$ be a current-frame token and $k_p$ the projected source-patch feature used by the sampler. With temperature $\tau=0.10$, we compute
\[
    a_{\ell p}=\operatorname{softmax}_{p}\!\left(
        \frac{\operatorname{norm}(q_\ell)^\top\operatorname{norm}(k_p)}{\tau}
    \right),
    \qquad
    m^{V}_p=\frac{1}{L}\sum_{\ell=1}^{L}a_{\ell p}.
\]
$m^V$ is the \emph{current-frame representation map}: it measures which source patches are most similar to the information carried by the sampled $v_t$ tokens.

For rendering, each per-view map is reshaped to its patch grid, resized to the corresponding input image with bicubic interpolation, and optionally smoothed with a Gaussian kernel of $\sigma=6$ pixels. The map is min--max normalized independently within each view and overlaid with opacity $0.45$ on the min--max normalized RGB frame with opacity $0.55$. We use the MAGMA colormap for $m^Z$ and VIRIDIS for $m^V$; color intensity is comparable within a map family and frame. Multi-camera panels preserve the model's camera order; for three-view RoboTwin observations the display uses the T-shaped canvas with the two wrist views above the front view.

\begin{figure}[H]
\centering
\includegraphics[width=0.36\textwidth]{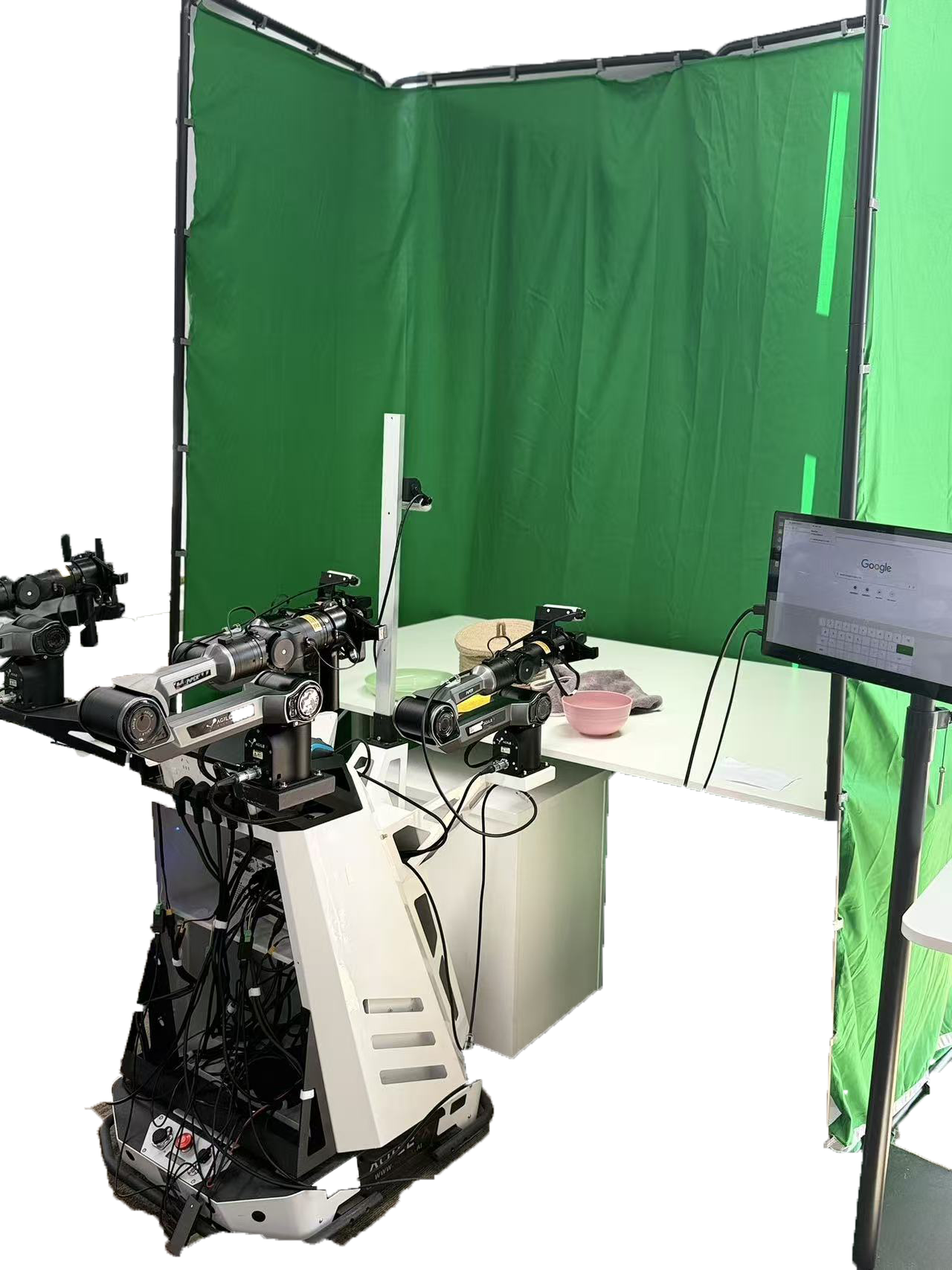}\hfill
\includegraphics[width=0.58\textwidth]{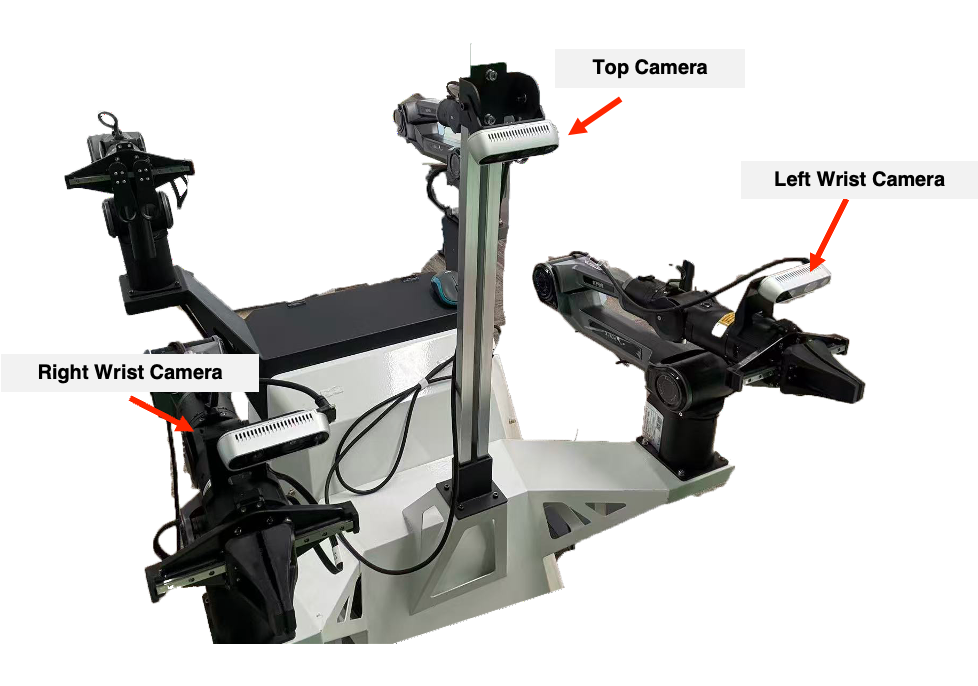}
\caption{Real-robot experimental platform. The AgileX Cobot is equipped with two Piper arms and standard grippers (left); three Intel Corp. RealSense D435 cameras provide one top and two wrist views (right).}
\label{fig:real_robot_platform}
\end{figure}

\section{Real-Robot Evaluation Details}
\label{sec:supp_real_robot_details}

\subsection{Experimental Platform}
\label{sec:supp_real_robot_platform}

Our experiments use an AgileX Cobot equipped with two 6-degree-of-freedom Piper arms and standard grippers (\cref{fig:real_robot_platform}). Robot communication, state acquisition, and joint-position command execution use the SDK provided by AgileX. Three Intel Corp. RealSense D435 cameras provide multi-view observations, of which the policy uses only the RGB streams. The action space comprises 14-D absolute joint positions (qpos). Teleoperated demonstrations are stored in the LeRobot dataset format~\citep{cadene2024lerobot}.

\subsection{Real-Time Chunking}
\label{sec:supp_rtc}

Evaluation scenes are produced by an automatic generator rather than placed by hand. We fit the distribution of object positions and orientations in the training set, sample from it with jitter, and hold the seeds fixed. Every policy therefore faces the same 50 initial scenes per task.

Copper-Policy uses real-time chunking (RTC)~\citep{black2025training} during training and deployment. During training, we simulate RTC delay by conditioning the Action Expert on a committed action prefix whose length is sampled from 0 to 6 steps. For real-robot deployment, asynchronous RTC executes the action chunk $a^+ := a_{t:t+H_a-1}$ in closed loop: while the robot executes a committed prefix, the policy server infers the remaining part of the next chunk from the latest multi-view observation. After an initial chunk is committed, each replanning step preserves a committed prefix whose length is determined by the estimated end-to-end inference delay, while the Action Expert regenerates the remaining, uncommitted portion; RTC then replans at 30\,Hz. This overlap between execution and inference maintains continuous action execution despite inference latency. The committed-prefix mechanism is specific to real-robot execution and is distinct from the action tokens the Action Expert otherwise generates during training and simulation evaluation.

\subsection{Protocol and Outcome Counts}
\label{sec:supp_real_robot}

All policies, including $\pi_{0.5}$ and Fast-WAM, follow the same evaluation protocol and scene setup. They are trained or fine-tuned on the same 300 demonstrations per task, deployed on the same platform with the same cameras and action space, and evaluated on 50 trials per task from shared initial-scene seeds. Table~\ref{tab:supp_real_robot_counts} reports full-success, half-success, and failure counts, scored as 1, 0.5, and 0. Trial outcomes are assigned by manual annotation of the recorded rollouts. A trial is a full success when the task goal is reached, a half success when the goal is only partially reached, and a failure otherwise. Half success is awarded only on towel folding, when the fold is completed but a visible crease or corner remains. Banana placement and bowl stacking admit no partial credit. The reported average uses all 150 trials per policy variant.

\begin{table}[H]
\centering
\scriptsize
\setlength{\tabcolsep}{5pt}
\renewcommand{\arraystretch}{0.90}
\begin{tabular}{l|ccc|c}
\toprule
\multirow{2}{*}{\raisebox{-4.5pt}{Variant}} & \multicolumn{3}{c|}{Full / half / failure counts} & \multirow{2}{*}{\raisebox{-4.5pt}{Avg.\ score}} \\
\cmidrule(lr){2-4}
& \shortstack[c]{Fold\\Towel} & \shortstack[c]{Put\\Banana} & \shortstack[c]{Stack\\Bowls} & \\
\midrule
$\pi_{0.5}$~\citep{intelligence2025pi_} & 47/1/2 & 47/0/3 & 47/0/3 & 94.3 \\
Fast-WAM~\citep{yuan2026fastwam} & 49/0/1 & 30/0/20 & 47/0/3 & 84.0 \\
\midrule
Copper-Policy & 44/5/1 & 48/0/2 & 50/0/0 & 96.3 \\
w/o World Input \& Future Expert & 41/2/7 & 41/0/9 & 44/0/6 & 84.7 \\
w/o Joint-Embedding Prediction & 1/2/47 & 13/0/37 & 50/0/0 & 43.3 \\
w/o Current-Frame Visual Features & 1/0/49 & 23/0/27 & 7/0/43 & 20.7 \\
w/o Joint-Embedding Prediction \& Visual Features & 0/0/50 & 2/0/48 & 0/0/50 & 1.3 \\
\bottomrule
\end{tabular}
\caption{Per-task outcome counts behind the aggregate real-robot scores.}
\label{tab:supp_real_robot_counts}
\end{table}

\paragraph{Additional qualitative observations.}
Successful Copper-Policy rollouts sometimes reach the task goals along trajectories visibly different from the teleoperated demonstrations. Fast-WAM often does not attempt the grasp in banana placement. In an additional unscored check outside the 150 reported trials, rotating the banana toward orientations present in the demonstrations recovered some grasp attempts.

%% file: main.bib
@article{yuan2026fastwam,
  title={Fast-WAM: Do World Action Models Need Test-time Future Imagination?},
  author={Tianyuan Yuan and Zibin Dong and Yicheng Liu and Hang Zhao},
  journal={arXiv preprint arXiv:2603.16666},
  year={2026},
  url={https://arxiv.org/abs/2603.16666}
}

@article{black2025training,
  title={Training-time action conditioning for efficient real-time chunking},
  author={Black, Kevin and Ren, Allen Z and Equi, Michael and Levine, Sergey},
  journal={arXiv preprint arXiv:2512.05964},
  year={2025}
}

@article{cai2026aha,
  title={AHA-WAM: Asynchronous Horizon-Adaptive World-Action Modeling with Observation-Guided Context Routing},
  author={Cai, Jisong and Ling, Long and Chu, Shiwei and Liu, Zhongshan and Kang, Jiayue and Liang, Zhixuan and Xu, Wenjie and Mao, Yinan and Zhang, Weinan and Yang, Xiaokang and others},
  journal={arXiv preprint arXiv:2606.09811},
  year={2026}
}

@article{spatialforcing2025,
  author    = {Li, Fuhao and Song, Wenxuan and Zhao, Han and Wang, Jingbo and Ding, Pengxiang and Wang, Donglin and Zeng, Long and Li, Haoang},
  title     = {Spatial Forcing: Implicit Spatial Representation Alignment For Vision-Language-Action Model},
  journal   = {arXiv preprint arXiv:2510.12276},
  year      = {2025},
}

@article{chen2025robotwin,
  title={Robotwin 2.0: A scalable data generator and benchmark with strong domain randomization for robust bimanual robotic manipulation},
  author={Chen, Tianxing and Chen, Zanxin and Chen, Baijun and Cai, Zijian and Liu, Yibin and Li, Zixuan and Liang, Qiwei and Lin, Xianliang and Ge, Yiheng and Gu, Zhenyu and others},
  journal={arXiv preprint arXiv:2506.18088},
  year={2025}
}

@article{intelligence2025pi_,
  title={{$\pi_{0.5}$}: A Vision-Language-Action Model with Open-World Generalization},
  author={{Physical Intelligence} and Black, Kevin and Brown, Noah and Darpinian, James and Dhabalia, Karan and Driess, Danny and Esmail, Adnan and Equi, Michael and Finn, Chelsea and Fusai, Niccolo and others},
  journal={arXiv preprint arXiv:2504.16054},
  year={2025}
}

@article{kim2026cosmos,
  title={Cosmos policy: Fine-tuning video models for visuomotor control and planning},
  author={Kim, Moo Jin and Gao, Yihuai and Lin, Tsung-Yi and Lin, Yen-Chen and Ge, Yunhao and Lam, Grace and Liang, Percy and Song, Shuran and Liu, Ming-Yu and Finn, Chelsea and others},
  journal={arXiv preprint arXiv:2601.16163},
  year={2026}
}

@inproceedings{zheng2026x,
  title={X-vla: Soft-prompted transformer as scalable cross-embodiment vision-language-action model},
  author={Zheng, Jinliang and Li, Jianxiong and Wang, Zhihao and Liu, Dongxiu and Kang, Xirui and Feng, Yuchun and Zheng, Yinan and Zou, Jiayin and Chen, Yilun and Zeng, Jia and others},
  booktitle={International Conference on Learning Representations},
  volume={2026},
  pages={60580--60606},
  year={2026}
}

@article{guo2026unified,
  title={Unified 4d world action modeling from video priors with asynchronous denoising},
  author={Guo, Jun and Li, Qiwei and Li, Peiyan and Chen, Zilong and Sun, Nan and Su, Yifei and Wang, Heyun and Zhang, Yuan and Li, Xinghang and Liu, Huaping},
  journal={arXiv preprint arXiv:2604.26694},
  year={2026}
}

@article{yang2026eventvla,
  title={EventVLA: Event-Driven Visual Evidence Memory for Long-Horizon Vision-Language-Action Policies},
  author={Yang, Ganlin and Tu, Zhangzheng and Yang, Yuqiang and Mao, Sitong and Dong, Junyi and Chen, Tianxing and Peng, Jiaqi and Xiong, Jing and Cao, Jiafei and Dai, Jifeng and others},
  journal={arXiv preprint arXiv:2606.20092},
  year={2026}
}

@article{ye2026starvla,
  title={{StarVLA-$\alpha$}: Reducing Complexity in Vision-Language-Action Systems},
  author={Ye, Jinhui and Gao, Ning and Yang, Senqiao and Zheng, Jinliang and Wang, Zixuan and Chen, Yuxin and Chen, Pengguang and Chen, Yilun and Liu, Shu and Jia, Jiaya},
  journal={arXiv preprint arXiv:2604.11757},
  year={2026}
}

@inproceedings{ICLR2025_49f80e4d,
 author = {Liu, Songming and Wu, Lingxuan and Li, Bangguo and Tan, Hengkai and Chen, Huayu and Wang, Zhengyi and Xu, Ke and Su, Hang and Zhu, Jun},
 booktitle = {International Conference on Learning Representations},
 editor = {Y. Yue and A. Garg and N. Peng and F. Sha and R. Yu},
 pages = {29982--30009},
 title = {RDT-1B: a Diffusion Foundation Model for Bimanual Manipulation},
 url = {https://proceedings.iclr.cc/paper_files/paper/2025/file/49f80e4d2471ad4f2edf4f5f1ab62339-Paper-Conference.pdf},
 volume = {2025},
 year = {2025}
}

@article{yang2026abot,
  title={Abot-m0: Vla foundation model for robotic manipulation with action manifold learning},
  author={Yang, Yandan and Zeng, Shuang and Lin, Tong and Chang, Xinyuan and Qi, Dekang and Xiao, Junjin and Liu, Haoyun and Chen, Ronghan and Chen, Yuzhi and Huo, Dongjie and others},
  journal={arXiv preprint arXiv:2602.11236},
  year={2026}
}

@article{cai2026xiaomi,
  title={Xiaomi-robotics-0: An open-sourced vision-language-action model with real-time execution},
  author={Cai, Rui and Guo, Jun and He, Xinze and Jin, Piaopiao and Li, Jie and Lin, Bingxuan and Liu, Futeng and Liu, Wei and Ma, Fei and Ma, Kun and others},
  journal={arXiv preprint arXiv:2602.12684},
  year={2026}
}

@INPROCEEDINGS{Ze-RSS-24, 
    AUTHOR    = {Yanjie Ze AND Gu Zhang AND Kangning Zhang AND Chenyuan Hu AND Muhan Wang AND Huazhe Xu}, 
    TITLE     = {{3D Diffusion Policy: Generalizable Visuomotor Policy Learning via Simple 3D Representations}}, 
    BOOKTITLE = {Proceedings of Robotics: Science and Systems}, 
    YEAR      = {2024}, 
    ADDRESS   = {Delft, Netherlands}, 
    MONTH     = {July}, 
    DOI       = {10.15607/RSS.2024.XX.067} 
}

@INPROCEEDINGS{Zhao-RSS-23, 
    AUTHOR    = {Tony Z. Zhao AND Vikash Kumar AND Sergey Levine AND Chelsea Finn}, 
    TITLE     = {{Learning Fine-Grained Bimanual Manipulation with Low-Cost Hardware}}, 
    BOOKTITLE = {Proceedings of Robotics: Science and Systems}, 
    YEAR      = {2023}, 
    ADDRESS   = {Daegu, Republic of Korea}, 
    MONTH     = {July}, 
    DOI       = {10.15607/RSS.2023.XIX.016} 
}

@article{zhang2026world,
  title={Do world action models generalize better than vlas? a robustness study},
  author={Zhang, Zhanguang and Li, Zhiyuan and Rahmati, Behnam and Yang, Rui Heng and Ma, Yintao and Rasouli, Amir and Pakdamansavoji, Sajjad and Wu, Yangzheng and Zhang, Lingfeng and Cao, Tongtong and others},
  journal={arXiv preprint arXiv:2603.22078},
  year={2026}
}

@article{li2026causal,
  title={Causal world modeling for robot control},
  author={Li, Lin and Zhang, Qihang and Luo, Yiming and Yang, Shuai and Wang, Ruilin and Han, Fei and Yu, Mingrui and Gao, Zelin and Xue, Nan and Zhu, Xing and others},
  journal={arXiv preprint arXiv:2601.21998},
  year={2026}
}

@InProceedings{pmlr-v305-zheng25a,
  title = 	 {FLARE: Robot Learning with Implicit World Modeling},
  author =       {Zheng, Ruijie and Wang, Jing and Reed, Scott and Bjorck, Johan and Fang, Yu and Hu, Fengyuan and Jang, Joel and Kundalia, Kaushil and Lin, Zongyu and Magne, Lo\"{\i}c and Narayan, Avnish and Tan, You Liang and Wang, Guanzhi and Wang, Qi and Xiang, Jiannan and Xu, Yinzhen and Ye, Seonghyeon and Kautz, Jan and Huang, Furong and Zhu, Yuke and Fan, Linxi},
  booktitle = 	 {Proceedings of The 9th Conference on Robot Learning},
  pages = 	 {3952--3971},
  year = 	 {2025},
  editor = 	 {Lim, Joseph and Song, Shuran and Park, Hae-Won},
  volume = 	 {305},
  series = 	 {Proceedings of Machine Learning Research},
  month = 	 {27--30 Sep},
  publisher =    {PMLR},
  url = 	 {https://proceedings.mlr.press/v305/zheng25a.html}
}

@inproceedings{sun2026vla,
  title={Vla-jepa: Enhancing vision-language-action model with latent world model},
  author={Sun, Jingwen and Zhang, Wenyao and Qi, Zekun and Ren, Shaojie and Liu, Zezhi and Zhu, Hanxin and Sun, Guangzhong and Jin, Xin and Chen, Zhibo},
  booktitle={European Conference on Computer Vision},
  pages={478--497},
  year={2026},
  organization={Springer}
}

@article{beingbeyond2026beingh07,
  title={{Being-H0.7}: A Latent World-Action Model from Egocentric Videos},
  author={Luo, Hao and Zhang, Wanpeng and Feng, Yicheng and Zheng, Sipeng and Xu, Haiweng and Xu, Chaoyi and Xi, Ziheng and Fu, Yuhui and Lu, Zongqing},
  journal={arXiv preprint arXiv:2605.00078},
  year={2026}
}

@article{su2026world,
  title={World guidance: World modeling in condition space for action generation},
  author={Su, Yue and Chen, Sijin and Shi, Haixin and Liu, Mingyu and Zhang, Zhengshen and Huang, Ningyuan and Zhong, Weiheng and Zhu, Zhengbang and Liu, Yuxiao and Liu, Xihui},
  journal={arXiv preprint arXiv:2602.22010},
  year={2026}
}

@InProceedings{pmlr-v267-zhou25t,
  title = 	 {{DINO}-{WM}: World Models on Pre-trained Visual Features enable Zero-shot Planning},
  author =       {Zhou, Gaoyue and Pan, Hengkai and LeCun, Yann and Pinto, Lerrel},
  booktitle = 	 {Proceedings of the 42nd International Conference on Machine Learning},
  pages = 	 {79115--79135},
  year = 	 {2025},
  editor = 	 {Singh, Aarti and Fazel, Maryam and Hsu, Daniel and Lacoste-Julien, Simon and Berkenkamp, Felix and Maharaj, Tegan and Wagstaff, Kiri and Zhu, Jerry},
  volume = 	 {267},
  series = 	 {Proceedings of Machine Learning Research},
  month = 	 {13--19 Jul},
  publisher =    {PMLR},
  url = 	 {https://proceedings.mlr.press/v267/zhou25t.html}
}

@article{maes2026leworldmodel,
  title={Leworldmodel: Stable end-to-end joint-embedding predictive architecture from pixels},
  author={Maes, Lucas and Lidec, Quentin Le and Scieur, Damien and LeCun, Yann and Balestriero, Randall},
  journal={arXiv preprint arXiv:2603.19312},
  year={2026}
}

@article{fei2025libero,
  title={Libero-plus: In-depth robustness analysis of vision-language-action models},
  author={Fei, Senyu and Wang, Siyin and Shi, Junhao and Dai, Zihao and Cai, Jikun and Qian, Pengfang and Ji, Li and He, Xinzhe and Zhang, Shiduo and Fei, Zhaoye and others},
  journal={arXiv preprint arXiv:2510.13626},
  year={2025}
}

@InProceedings{Assran_2023_CVPR,
    author    = {Assran, Mahmoud and Duval, Quentin and Misra, Ishan and Bojanowski, Piotr and Vincent, Pascal and Rabbat, Michael and LeCun, Yann and Ballas, Nicolas},
    title     = {Self-Supervised Learning From Images With a Joint-Embedding Predictive Architecture},
    booktitle = {Proceedings of the IEEE/CVF Conference on Computer Vision and Pattern Recognition (CVPR)},
    month     = {June},
    year      = {2023},
    pages     = {15619-15629}
}

@article{bardes2024revisiting,
  title={Revisiting feature prediction for learning visual representations from video},
  author={Bardes, Adrien and Garrido, Quentin and Ponce, Jean and Chen, Xinlei and Rabbat, Michael and LeCun, Yann and Assran, Mahmoud and Ballas, Nicolas},
  journal={arXiv preprint arXiv:2404.08471},
  year={2024}
}

@article{grill2020bootstrap,
  title={Bootstrap your own latent-a new approach to self-supervised learning},
  author={Grill, Jean-Bastien and Strub, Florian and Altch{\'e}, Florent and Tallec, Corentin and Richemond, Pierre and Buchatskaya, Elena and Doersch, Carl and Avila Pires, Bernardo and Guo, Zhaohan and Gheshlaghi Azar, Mohammad and others},
  journal={Advances in neural information processing systems},
  volume={33},
  pages={21271--21284},
  year={2020}
}

@article{zhang2026learning,
  title={Learning 4D Geometric Priors for Inference-Efficient World Action Models},
  author={Zhang, Jianjun and Zhu, Jian and Su, Taiyi and Ma, Chong and Huang, Zitai and Xu, Yi and Wang, Hanli},
  journal={arXiv preprint arXiv:2607.05468},
  year={2026}
}

@article{zhao2026rynnworld,
  title={RynnWorld-4D: 4D Embodied World Models for Robotic Manipulation},
  author={Zhao, Haoyu and Zhao, Xingyue and Huang, Siteng and Li, Xin and Zhao, Deli and Li, Zhongyu},
  journal={arXiv preprint arXiv:2607.06559},
  year={2026}
}

@inproceedings{NEURIPS2023_8c3c6668,
 author = {Liu, Bo and Zhu, Yifeng and Gao, Chongkai and Feng, Yihao and Liu, Qiang and Zhu, Yuke and Stone, Peter},
 booktitle = {Advances in Neural Information Processing Systems},
 doi = {10.52202/075280-1939},
 editor = {A. Oh and T. Naumann and A. Globerson and K. Saenko and M. Hardt and S. Levine},
 pages = {44776--44791},
 publisher = {Curran Associates, Inc.},
 title = {LIBERO: Benchmarking Knowledge Transfer for Lifelong Robot Learning},
 url = {https://proceedings.neurips.cc/paper_files/paper/2023/file/8c3c666820ea055a77726d66fc7d447f-Paper-Datasets_and_Benchmarks.pdf},
 volume = {36},
 year = {2023}
}

@inproceedings{
feng2026see,
title={See What Matters: Differentiable Grid Sample Pruning for Generalizable Vision-Language-Action Model},
author={Yixu Feng and Zinan Zhao and Yanxiang Ma and Chenghao Xia and Chengbin Du and Yunke Wang and Chang Xu},
booktitle={Forty-third International Conference on Machine Learning},
year={2026},
url={https://openreview.net/forum?id=HpIhfqaLwa}
}

@inproceedings{
lipman2023flow,
title={Flow Matching for Generative Modeling},
author={Yaron Lipman and Ricky T. Q. Chen and Heli Ben-Hamu and Maximilian Nickel and Matthew Le},
booktitle={The Eleventh International Conference on Learning Representations },
year={2023},
url={https://openreview.net/forum?id=PqvMRDCJT9t}
}

@article{
  liang2025mixtureoftransformers,
  title={Mixture-of-Transformers: A Sparse and Scalable Architecture for Multi-Modal Foundation Models},
  author={Weixin Liang and LILI YU and Liang Luo and Srini Iyer and Ning Dong and Chunting Zhou and Gargi Ghosh and Mike Lewis and Wen-tau Yih and Luke Zettlemoyer and Xi Victoria Lin},
  journal={Transactions on Machine Learning Research},
  issn={2835-8856},
  year={2025},
  url={https://openreview.net/forum?id=Nu6N69i8SB},
  note={}
}

@incollection{ha2018world,
  title = {Recurrent World Models Facilitate Policy Evolution},
  author = {Ha, David and Schmidhuber, J{\"u}rgen},
  booktitle = {Advances in Neural Information Processing Systems 31},
  pages = {2451--2463},
  year = {2018},
  publisher = {Curran Associates, Inc.},
  url = {https://papers.nips.cc/paper/7512-recurrent-world-models-facilitate-policy-evolution},
  note = "\url{https://worldmodels.github.io}",
}

@inproceedings{hafner2019learning,
  title={Learning latent dynamics for planning from pixels},
  author={Hafner, Danijar and Lillicrap, Timothy and Fischer, Ian and Villegas, Ruben and Ha, David and Lee, Honglak and Davidson, James},
  booktitle={International conference on machine learning},
  pages={2555--2565},
  year={2019},
  organization={PMLR}
}

@Article{Hafner2025,
author={Hafner, Danijar
and Pasukonis, Jurgis
and Ba, Jimmy
and Lillicrap, Timothy},
title={Mastering diverse control tasks through world models},
journal={Nature},
year={2025},
month={Apr},
day={01},
volume={640},
number={8059},
pages={647-653},
issn={1476-4687},
doi={10.1038/s41586-025-08744-2},
url={https://doi.org/10.1038/s41586-025-08744-2}
}

@InProceedings{pmlr-v205-nair23a,
  title = 	 {R3M: A Universal Visual Representation for Robot Manipulation},
  author =       {Nair, Suraj and Rajeswaran, Aravind and Kumar, Vikash and Finn, Chelsea and Gupta, Abhinav},
  booktitle = 	 {Proceedings of The 6th Conference on Robot Learning},
  pages = 	 {892--909},
  year = 	 {2023},
  editor = 	 {Liu, Karen and Kulic, Dana and Ichnowski, Jeff},
  volume = 	 {205},
  series = 	 {Proceedings of Machine Learning Research},
  month = 	 {14--18 Dec},
  publisher =    {PMLR},
  url = 	 {https://proceedings.mlr.press/v205/nair23a.html}
}

@inproceedings{
ma2023vip,
title={{VIP}: Towards Universal Visual Reward and Representation via Value-Implicit Pre-Training},
author={Yecheng Jason Ma and Shagun Sodhani and Dinesh Jayaraman and Osbert Bastani and Vikash Kumar and Amy Zhang},
booktitle={The Eleventh International Conference on Learning Representations },
year={2023},
url={https://openreview.net/forum?id=YJ7o2wetJ2}
}

@InProceedings{pmlr-v205-radosavovic23a,
  title = 	 {Real-World Robot Learning with Masked Visual Pre-training},
  author =       {Radosavovic, Ilija and Xiao, Tete and James, Stephen and Abbeel, Pieter and Malik, Jitendra and Darrell, Trevor},
  booktitle = 	 {Proceedings of The 6th Conference on Robot Learning},
  pages = 	 {416--426},
  year = 	 {2023},
  editor = 	 {Liu, Karen and Kulic, Dana and Ichnowski, Jeff},
  volume = 	 {205},
  series = 	 {Proceedings of Machine Learning Research},
  month = 	 {14--18 Dec},
  publisher =    {PMLR},
  url = 	 {https://proceedings.mlr.press/v205/radosavovic23a.html}
}

@INPROCEEDINGS{Karamcheti-RSS-23, 
    AUTHOR    = {Siddharth Karamcheti AND Suraj Nair AND Annie S Chen AND Thomas Kollar AND Chelsea Finn AND Dorsa Sadigh AND Percy Liang}, 
    TITLE     = {{Language-Driven Representation Learning for Robotics}}, 
    BOOKTITLE = {Proceedings of Robotics: Science and Systems}, 
    YEAR      = {2023}, 
    ADDRESS   = {Daegu, Republic of Korea}, 
    MONTH     = {July}, 
    DOI       = {10.15607/RSS.2023.XIX.032} 
}

@inproceedings{ICLR2025_45d74e19,
 author = {Ye, Seonghyeon and Jang, Joel and Jeon, Byeongguk and Joo, Se June and Yang, Jianwei and Peng, Baolin and Mandlekar, Ajay and Tan, Reuben and Chao, Yu-Wei and Lin, Bill Yuchen and Liden, Lars and Lee, Kimin  and Gao, Jianfeng and Zettlemoyer, Luke and Fox, Dieter and Seo, Minjoon},
 booktitle = {International Conference on Learning Representations},
 editor = {Y. Yue and A. Garg and N. Peng and F. Sha and R. Yu},
 pages = {28213--28239},
 title = {Latent Action Pretraining from Videos},
 url = {https://proceedings.iclr.cc/paper_files/paper/2025/file/45d74e190008c7bff2845ffc8e3facd3-Paper-Conference.pdf},
 volume = {2025},
 year = {2025}
}

@InProceedings{pmlr-v235-bruce24a,
  title = 	 {Genie: Generative Interactive Environments},
  author =       {Bruce, Jake and Dennis, Michael D and Edwards, Ashley and Parker-Holder, Jack and Shi, Yuge and Hughes, Edward and Lai, Matthew and Mavalankar, Aditi and Steigerwald, Richie and Apps, Chris and Aytar, Yusuf and Bechtle, Sarah Maria Elisabeth and Behbahani, Feryal and Chan, Stephanie C.Y. and Heess, Nicolas and Gonzalez, Lucy and Osindero, Simon and Ozair, Sherjil and Reed, Scott and Zhang, Jingwei and Zolna, Konrad and Clune, Jeff and Freitas, Nando De and Singh, Satinder and Rockt\"{a}schel, Tim},
  booktitle = 	 {Proceedings of the 41st International Conference on Machine Learning},
  pages = 	 {4603--4623},
  year = 	 {2024},
  editor = 	 {Salakhutdinov, Ruslan and Kolter, Zico and Heller, Katherine and Weller, Adrian and Oliver, Nuria and Scarlett, Jonathan and Berkenkamp, Felix},
  volume = 	 {235},
  series = 	 {Proceedings of Machine Learning Research},
  month = 	 {21--27 Jul},
  publisher =    {PMLR},
  url = 	 {https://proceedings.mlr.press/v235/bruce24a.html}
}

@InProceedings{pmlr-v270-kim25c,
  title = 	 {OpenVLA: An Open-Source Vision-Language-Action Model},
  author =       {Kim, Moo Jin and Pertsch, Karl and Karamcheti, Siddharth and Xiao, Ted and Balakrishna, Ashwin and Nair, Suraj and Rafailov, Rafael and Foster, Ethan P and Sanketi, Pannag R and Vuong, Quan and Kollar, Thomas and Burchfiel, Benjamin and Tedrake, Russ and Sadigh, Dorsa and Levine, Sergey and Liang, Percy and Finn, Chelsea},
  booktitle = 	 {Proceedings of The 8th Conference on Robot Learning},
  pages = 	 {2679--2713},
  year = 	 {2025},
  editor = 	 {Agrawal, Pulkit and Kroemer, Oliver and Burgard, Wolfram},
  volume = 	 {270},
  series = 	 {Proceedings of Machine Learning Research},
  month = 	 {06--09 Nov},
  publisher =    {PMLR},
  url = 	 {https://proceedings.mlr.press/v270/kim25c.html}
}

@article{black2024pi_0,
  title={{$\pi_0$}: A Vision-Language-Action Flow Model for General Robot Control},
  author={Black, Kevin and Brown, Noah and Driess, Danny and Esmail, Adnan and Equi, Michael and Finn, Chelsea and Fusai, Niccolo and Groom, Lachy and Hausman, Karol and Ichter, Brian and others},
  journal={arXiv preprint arXiv:2410.24164},
  year={2024}
}

@INPROCEEDINGS{Chi-RSS-23, 
    AUTHOR    = {Cheng Chi AND Siyuan Feng AND Yilun Du AND Zhenjia Xu AND Eric Cousineau AND Benjamin CM Burchfiel AND Shuran Song}, 
    TITLE     = {{Diffusion Policy: Visuomotor Policy Learning via Action Diffusion}}, 
    BOOKTITLE = {Proceedings of Robotics: Science and Systems}, 
    YEAR      = {2023}, 
    ADDRESS   = {Daegu, Republic of Korea}, 
    MONTH     = {July}, 
    DOI       = {10.15607/RSS.2023.XIX.026} 
}

@inproceedings{NEURIPS2023_1d5b9233,
 author = {Du, Yilun and Yang, Sherry and Dai, Bo and Dai, Hanjun and Nachum, Ofir and Tenenbaum, Josh and Schuurmans, Dale and Abbeel, Pieter},
 booktitle = {Advances in Neural Information Processing Systems},
 doi = {10.52202/075280-0403},
 editor = {A. Oh and T. Naumann and A. Globerson and K. Saenko and M. Hardt and S. Levine},
 pages = {9156--9172},
 publisher = {Curran Associates, Inc.},
 title = {Learning Universal Policies via Text-Guided Video Generation},
 url = {https://proceedings.neurips.cc/paper_files/paper/2023/file/1d5b9233ad716a43be5c0d3023cb82d0-Paper-Conference.pdf},
 volume = {36},
 year = {2023}
}

@inproceedings{ICLR2024_b2d4051f,
 author = {Ko, Po-Chen and Mao, Jiayuan and Du, Yilun and Sun, Shao-Hua and Tenenbaum, Joshua B},
 booktitle = {International Conference on Learning Representations},
 editor = {B. Kim and Y. Yue and S. Chaudhuri and K. Fragkiadaki and M. Khan and Y. Sun},
 pages = {40938--40958},
 title = {Learning to Act from Actionless Videos through Dense Correspondences},
 url = {https://proceedings.iclr.cc/paper_files/paper/2024/file/b2d4051f03a7038a2771dfbbe5c7b54e-Paper-Conference.pdf},
 volume = {2024},
 year = {2024}
}

@article{cheang2024gr,
  title={Gr-2: A generative video-language-action model with web-scale knowledge for robot manipulation},
  author={Cheang, Chi-Lam and Chen, Guangzeng and Jing, Ya and Kong, Tao and Li, Hang and Li, Yifeng and Liu, Yuxiao and Wu, Hongtao and Xu, Jiafeng and Yang, Yichu and others},
  journal={arXiv preprint arXiv:2410.06158},
  year={2024}
}

@misc{liu2026stagewamjointembeddingstageprediction,
      title={StageWAM: Joint-Embedding Stage Prediction for World-Action Models in Robot Manipulation}, 
      author={Xiao Liu and Yuguang Yang and Xi Wang and Kai Jiang and Cheng Chi and Yong Xu and Wenchao Ding and Yilun Chen and Yan Wang},
      year={2026},
      eprint={2608.10780},
      archivePrefix={arXiv},
      primaryClass={cs.RO},
      url={https://arxiv.org/abs/2608.10780}, 
}

@misc{lin2026jepawamlearningvisionlanguageactionpolicies,
      title={JEPA-WAM: Learning Vision-Language-Action Policies with Joint-Embedding World Modeling}, 
      author={Yihan Lin and Jiawei He and Shifeng Bao and Chen Zhao and Yang Li and Xiaobo Wang and Yan Wang and Cheng Chi and Jing Zhang},
      year={2026},
      eprint={2608.09381},
      archivePrefix={arXiv},
      primaryClass={cs.RO},
      url={https://arxiv.org/abs/2608.09381}, 
}

@misc{yang2026lilawamlightweightlatentreasoning,
      title={LiLa-WAM: Lightweight Latent Reasoning World-Action Model for Robotic Manipulation}, 
      author={Fan Yang and Yuting Su and Xiaobo Wang and Yuncheng You and Fugui Fan and Yuting Wu and Minghui Wu and Chenxu Zhao and JiaHong Ning and Peiguang Jing},
      year={2026},
      eprint={2608.03701},
      archivePrefix={arXiv},
      primaryClass={cs.RO},
      url={https://arxiv.org/abs/2608.03701}, 
}

@misc{liu2026g05autoregressivestreamrobot,
      title={G0.5: One Autoregressive Stream for Robot Reasoning and Action}, 
      author={Yicheng Liu and Zibin Dong and Baijun Ye and Tianyuan Yuan and Tao Jiang and Anqi Yang and Shicheng Cao and Haonan Liu and Yue Sun and Zihan Guo and Xiao Liu and Dong Ke and Changxun Pan and Chenru Wu and Tailai Cheng and Xiaoshu Ren and Xinlei Zhang and Jianning Cui and Zijie Zhao and Haoyu Zhang and Kaiming Xu and Haodong Yang and Bowen Zhang and Jiahui Niu and Shaoting Zhu and Shiduo Zhang and Hang Zhao},
      year={2026},
      eprint={2608.11739},
      archivePrefix={arXiv},
      primaryClass={cs.RO},
      url={https://arxiv.org/abs/2608.11739}, 
}

@misc{cadene2024lerobot,
    author = {Cadene, Remi and Alibert, Simon and Soare, Alexander and Gallouedec, Quentin and Zouitine, Adil and Palma, Steven and Kooijmans, Pepijn and Aractingi, Michel and Shukor, Mustafa and Aubakirova, Dana and Russi, Martino and Capuano, Francesco and Pascal, Caroline and Choghari, Jade and Meftah, Khalil and Ellerbach, Maxime and Moss, Jess and Wolf, Thomas},
    title = {LeRobot: State-of-the-art Machine Learning for Real-World Robotics in Pytorch},
    howpublished = "\url{https://github.com/huggingface/lerobot}",
    year = {2024}
}
